\RequirePackage{fix-cm}
\documentclass[twocolumn]{svjour3}          
\smartqed  
\usepackage{graphicx}
\usepackage{color}
\usepackage{xcolor}
\newcommand{\smallersec}[1]{\paragraph{\bf #1:}}
\usepackage{enumitem}
\newcommand\obj[1]{\texttt{#1}}
\usepackage{tabularx}
\newcommand{\smallsec}[1]{\paragraph {\bf #1:}}
\usepackage{subcaption}
\usepackage{booktabs}
\usepackage{hyperref}
\usepackage{wrapfig}
\newcommand{\ra}[1]{\renewcommand{\arraystretch}{#1}}
\usepackage{breakcites}

\usepackage[citestyle=authoryear, style=apa]{biblatex}
\let\oldcite\cite
\renewcommand*\cite[1]{(\oldcite{#1})}

\begin{document}

\title{REVISE: A Tool for Measuring and Mitigating Bias in Visual Datasets
}


\author{Angelina Wang \and
        Alexander Liu \and
        Ryan Zhang \and
        Anat Kleiman \and
        Leslie Kim \and
        Dora Zhao \and
        Iroha Shirai \and
        Arvind Narayanan \and
        Olga Russakovsky
}


\institute{Angelina Wang \at
              \email{angelina.wang@princeton.edu}           
}

\date{}

\maketitle

\begin{abstract}
Machine learning models are known to perpetuate and even amplify the biases present in the data. However, these data biases frequently do not become apparent until after the models are deployed. Our work tackles this issue and  enables the preemptive analysis of large-scale datasets. REVISE (REvealing VIsual biaSEs) is a tool that assists in the investigation of a visual dataset, surfacing potential biases  along three dimensions: (1) object-based, (2) person-based, and (3) geography-based. Object-based biases relate to the size, context, or diversity of the depicted objects. Person-based metrics focus on analyzing the portrayal of people within the dataset. Geography-based analyses consider the representation of different geographic locations. These three dimensions are deeply intertwined in how they interact to bias a dataset, and REVISE sheds light on this; the responsibility then lies with the user to consider the cultural and historical context, and to determine which of the revealed biases may be problematic. The tool further assists the user by suggesting actionable steps that may be taken to mitigate the revealed biases. Overall, the key aim of our work is to tackle the machine learning bias problem early in the pipeline.  REVISE is available at \url{https://github.com/princetonvisualai/revise-tool}.
\end{abstract}

\section{Introduction}
\begin{figure}[t]
\centering
\includegraphics[height=2.7cm]{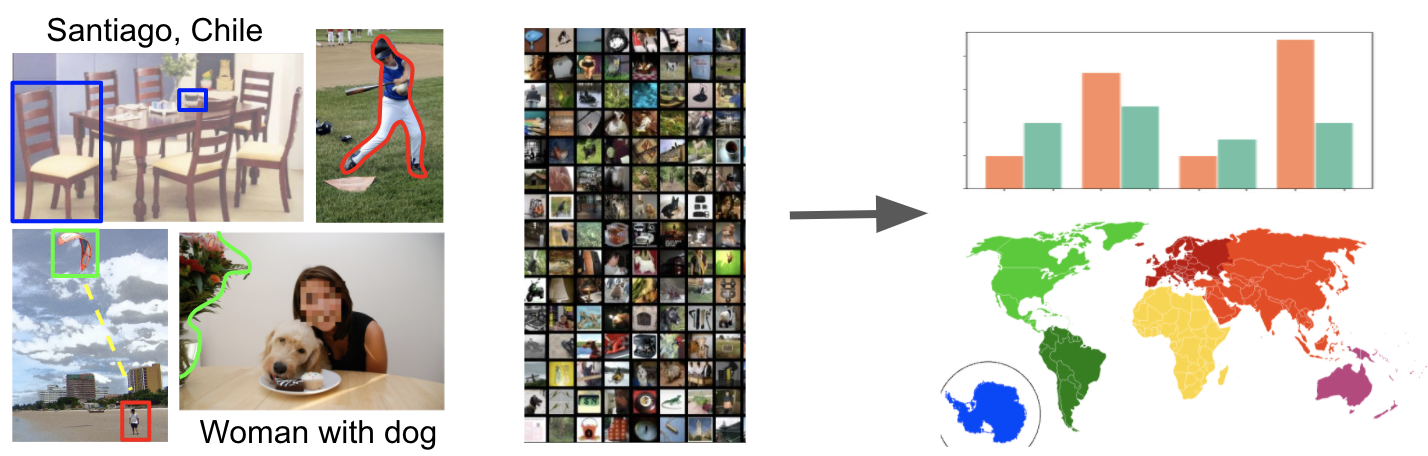}
\caption{Our tool takes in as input a visual dataset and its annotations, and outputs metrics, seeking to produce insights and possible actions.}
\label{fig:sum}
\end{figure}

Computer vision dataset bias is a well-known and much-studied problem. \textcite{Torralba11} highlighted the fact that every dataset is a unique slice through the visual world, representing a particular distribution of visual data. Since then, researchers have noted the under-representation of object classes~\cite{imbalance, Ouyang16, Yang14, Salakhutdinov11, Buda17, Liu09}, object contexts~\cite{Choi12, Rosenfeld18}, object sub-categories~\cite{Zhu14}, scenes~\cite{zhou2017places}, gender~\cite{Kay15}, gender contexts~\cite{Zhao17, Burns18}, skin tones~\cite{Buolamwini18, Wilson19}, geographic locations~\cite{Shankar17} and cultures~\cite{Devries19}. The downstream effects of these under-representations range from the more innocuous, like limited generalization of car classifiers~\cite{Torralba11}, to the much more serious, like having deep societal implications in automated facial analysis~\cite{Buolamwini18, Hill20}. Efforts such as Datasheets for Datasets~\cite{Gebru18} have played an important role in encouraging dataset transparency through articulating the intent of the dataset creators, summarizing the data collection processes, and warning downstream dataset users of potential biases in the data. However, this alone is not sufficient, as there is no algorithm to identify all biases hiding in the data, and manual review is not a feasible strategy given the scale of modern datasets.

\smallersec{Bias detection tool} To mitigate this issue, we provide an automated tool for REvealing VIsual biaSEs (REVISE) in datasets.  REVISE is a broad-purpose tool for surfacing the under- and different- representations hiding within visual datasets.  For the current exploration we limit ourselves to three sets of metrics: (1) object-based, (2) person-based and (3) geography-based.

Object-based analysis is most familiar to the computer vision community~\cite{Torralba11}, as many of the popular visual recognition datasets are object-centric~\cite{Everingham10,imagenet}. Thus, these analyses focus on considering statistics about object frequency, scale, context, or diversity of representation.

Person-based analyses began to gain attention with research showing unequal performance for people of different genders and skin tones~\cite{Gebru18,Zhao17}. This line of analysis considers the representation of people of different demographics within the dataset, and allows the user to assess what potential downstream consequences this may have in order to consider how best to intervene. It also builds on the object-based analysis by considering how the representation of objects with people of different demographic groups differs.

Finally, geography-based analysis considers the portrayal of different geographic regions within the dataset; this is a relatively new but very important conversation within the community~\cite{Shankar17, Devries19}. This axis of analysis is deeply intertwined with the previous two, as geography influences both the types of objects that are represented, as well as the different people that are pictured.

We imagine two primary use cases for our tool: (1) dataset builders can use the actionable insights produced by our tool during the process of dataset compilation to guide the direction of further data collection, and (2)  dataset users who train models can use the tool to understand what kinds of biases their models may inherit as a result of training on a particular dataset. 

\smallersec{Example Findings} REVISE automatically surfaces a variety of metrics that highlight unrepresentative or anomalous patterns in the dataset. To validate the usefulness of the tool, we have used it to analyze several datasets commonly used in computer vision: COCO~\cite{Lin14}, OpenImages~\cite{openimages}, 

\noindent YFCC100m \cite{Thomee16}, and BDD100K~\cite{yu2020bdd}. Some examples of the kinds of automatic insights our tool has found include:
\begin{itemize}
\item In the object detection dataset COCO~\cite{Lin14}, some objects, e.g., \obj{airplane}, \obj{bed} and \obj{pizza}, are frequently large in the image. This is because fewer images of \obj{airplanes} appear in the sky (far away; small) than on the ground (close-up; large). This may be a problem since object size plays a key role in recognition accuracy. One mitigation is to query for images of \obj{airplane} appearing in scenes of \obj{mountains, desert, sky}.
\item The OpenImages dataset~\cite{openimages} depicts a large number of people who are too small in the image for human annotators to determine their gender; nevertheless, we found that annotators infer that they are \obj{male} 69\% of the time, especially in scenes of \obj{outdoor sports fields, parks}. Computer vision researchers might want to exercise caution with these gender annotations so they don't propagate into the model.
\item In the computer vision and multimedia dataset 

YFCC100m (Yahoo Flickr Creative Commons 100 million)~\cite{Thomee16} images come from 196 different countries. However, we estimate that for around 47\% of those countries --- especially in developing regions of the world --- the images are predominantly photos taken by visitors to the country rather than by locals, potentially resulting in a stereotypical portrayal.
\end{itemize}

A benefit of our tool is that a user doesn't need to have specific biases in mind, as these can be hard to enumerate. Rather, the tool automatically surfaces unusual patterns. REVISE cannot automatically say which of these patterns, or lack of patterns, are problematic, and leaves that analysis up to the user's judgment and expertise. We note that ``bias" is a contested term, and while our tool seeks to surface a variety of findings that are interesting to dataset creators and users, not all may be considered forms of bias by everyone.

\section{Related Work}

\smallersec{Data collection} Visual datasets are constructed in various ways, with the most common being through keyword queries to search engines, whether singular (e.g., ImageNet~\cite{imagenet}) or pairwise (e.g., COCO~\cite{Lin14}), or by scraping websites like Flickr (e.g., YFCC100m~\cite{Thomee16}, OpenImages~\cite{openimages}). There is extensive preprocessing and cleaning done on the datasets. Human annotators, sometimes in conjunction with automated tools~\cite{zhou2017places}, then assign various labels and annotations. Dataset collectors put in significant effort to deal with problems like long-tails to ensure a more balanced distribution, and intra-class diversity by doing things like explicitly seeking out non-iconic images beyond just the object itself in focus.

\smallersec{Dataset Bias} Rather than pick a single definition, we adopt an inclusive notion of bias and seek to highlight ways in which the dataset builder can monitor and control the distribution of their data.
Proposed ways to deal with dataset bias include cross-dataset analysis~\cite{Torralba11} and having the machine learning community learn from data collection approaches of other disciplines~\cite{Jo20, Brown14}. Recent work~\cite{Prabhu20} has looked at dataset issues related to consent and justice; the authors advocate for enforcing Institutional Review Board (IRB) approval for large scale datasets. Although we have limited the scope of our work to the contents of the dataset itself, there are much broader questions of fairness to be considered regarding the role that datasets play~\cite{denton2020contest, paullada2020discontents}. Constructive solutions will need to combine automated analysis with human judgement as automated methods cannot yet understand things like the historical context that led to an observed statistical imbalance in the dataset. Our work takes this approach by automatically supplying a host of new metrics for analyzing a dataset along with actions that can be taken to mitigate these findings. However, the responsibility lies with the user to select next steps. The tool is open-source, lowering the resource and effort barrier to creating ethical datasets~\cite{Jo20}.

\smallersec{Computer vision tools} \textcite{Hoiem12} built a tool to diagnose the weaknesses of object detector models in order to help improve them. More recently, tools in the video domain \cite{Alwassel18} are looking into forms of dataset bias in activity recognition \cite{Sigurdsson17}. We similarly in spirit hope to build a tool that will, as one goal, help dataset curators be aware of the patterns and biases present in their datasets so they can iteratively make adjustments.

\smallersec{Algorithmic fairness} In addition to looking at how models trained on one dataset generalize poorly to others~\cite{Tommasi15, Torralba11}, many more forms of dataset bias are being increasingly noticed in the fairness domain~\cite{Caliskan17, Mehrabi19, Yang20}. There has been significant work looking at how to deal with this from the algorithm side~\cite{Dwork12, khosla12, Dwork17, Wang20} with varying definitions of fairness~\cite{Kilbertus17, Zhang18, Pleiss17, Gajane17, Srebo16} that are often deemed to be mathematically incompatible with each other~\cite{Chouldechova17, Kleinberg17}, but in this work, we look at the problem earlier in the pipeline from the dataset side.

\smallersec{Automated bias detectors}
Facebook's Fairness Flow \cite{fb2021flow} focuses on assessing model predictions for their contextual fairness, though also looks at the labels themselves. IBM's AI Fairness 360 \cite{Bellamy18} similarly discovers biases in machine learning models, and also looks into the datasets. However, its look into dataset biases is limited in that it first trains a model on that dataset, then interrogates this trained model with specific questions. 
On the other hand, REVISE looks directly at the dataset and its annotations to discover model-agnostic patterns. The Dataset Nutrition Label \cite{Holland18} is a recent project that assesses machine learning datasets. Differently, our approach works on visual rather than tabular data which allows us to use additional computer vision techniques, and goes deeper in terms of presenting a variety of graphs and statistical results. \textcite{Swinger19} look at automatic detection of biases in word embeddings, but we look at patterns in visual images and their annotations. Amazon SageMaker Clarify~\cite{amazon2021clarify} also works to detect bias in training data, but only along the person-based axis, and not object nor geography. Similarly, Google's Know Your Data~\cite{pair2021kyd} also aims to help mitigate bias issues in image datasets. However, their tool currently only works on TensorFlow image datasets, whereas REVISE will work for any local image dataset. This has the benefit of allowing dataset creators to iteratively query our tool during the development process of their dataset, as well as dataset users to apply it to a private or proprietary dataset.

\section{Tool Overview}
\label{sec:tool}
REVISE is a general tool intended to yield insights at varying levels of granularity. As an input, it simply requires an image dataset and any available annotations. The tool has the ability to fully automatically compute a host of metrics, to be described in Sec.~\ref{sec:obj_metrics}, \ref{sec:gen_metrics}, and \ref{sec:geo_metrics}, broken down by the axes of object, person, and geography. Which metrics can be computed depends on the annotations available, e.g., gender labels are required to compute statistics about different gender representations. To perform analyses beyond just the annotations provided, we also use external tools and pretrained models, such as Fasttext language detection~\cite{joulin2016bag, joulin2016fasttext}, Places scene detection~\cite{zhou2017places}, and automatic feature extraction~\cite{resnetgithub} to derive some of our metrics, and acknowledge these models themselves may contain biases. The metrics are often situated to provide a user with anomalous patterns, such as when the size distribution of an object class is highly non-uniform, and correspondingly provide automatic data-driven insights on how one might correct for this distribution. However, a metric itself has no normative claim on its own; ultimately it is up to the user to determine whether the automatically-surfaced patterns deviate enough from an intended distribution that this would be a problem for the downstream application of models trained on the dataset.

\begin{figure}
\centering
\includegraphics[width=0.48\textwidth]{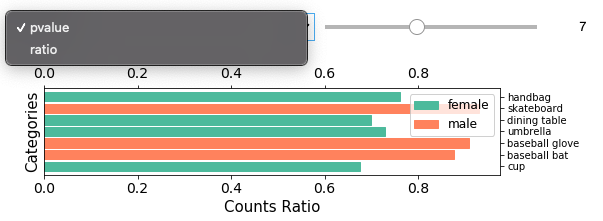}
\caption{Example interface of a metric in our notebook. A dropdown menus allow for sorting by p-value or ratio, and a sliding bar allows adjusting the number of examples shown in the graph. Best viewed in color.}
\label{fig:notebook}
\end{figure}

\begin{figure}
\centering
\includegraphics[width=0.49\textwidth]{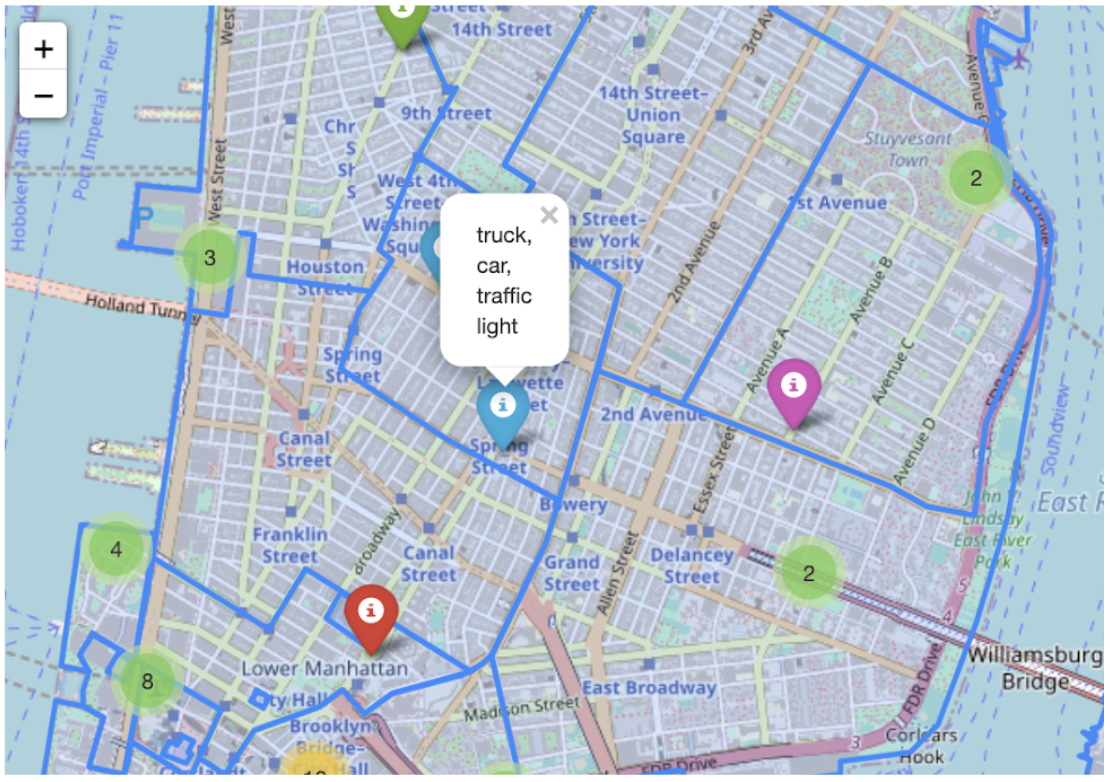}
\caption{Interface for exploring datasets with geography annotations. Interactive features allow viewing both image distribution by geography, as well as a bubble showing the labels of a specific image.
}
\label{fig:bdd_vis}
\end{figure}

\smallsec{Tool} Practically, REVISE takes the form of a Jupyter notebook interface that allows interactive exploration, as shown in Figs.~\ref{fig:notebook} and~\ref{fig:bdd_vis}. For privacy reasons, all analyses are run on a user's local machine. By default, the code to compute the metrics are largely abstracted away. However, all code is open-sourced such that a user can perform any personal customization to the metrics to fit their intended use-case. After multiple rounds of iterations with potential users, we have added a number of features to increase adoptability and usability of the tool. For one, we have created a video that will demonstrate to potential users what the interface of the tool looks like, and allow them to get a sense of whether this would be useful for their purposes. We have also included a feature that automatically generates a summary PDF as a result of running the tool and exploring the notebook. This supplements the dynamic nature of the tool with a static component that summarizes the key findings. Additionally, the biggest hurdle for a user to get started with their own dataset is the process of setting their dataset up to be in the standardized format that our tool requires. To this end, we have created a comprehensive testing script  that provides informative feedback to ensure a user has inputted their dataset in the proper format before it is run through the tool.

\smallsec{Axes of analyses} The analyses that can ultimately be performed depend on the annotations available:

\begin{enumerate}
    \item {\bf Object-based insights} require instance labels and, if available, their corresponding bounding boxes and object category. Datasets are frequently collected together with manual annotations, but we also use automated computer vision techniques to infer some semantic labels, like scenes.
    \item {\bf Person-based insights} require sensitive attribute labels of the people in the images. The tool is general enough that given labels of any grouping of people, such as racial groups, the corresponding analyses can be performed. If the attribute labels are ordinal, such as quantized age or skin tone, additional regression analyses are available.
    \item{\bf Geography-based insights} flexibly allow for labels in two possible formats: 1) region labels as strings, e.g., ``Portugal", ``Nigeria", or 2) GPS latitude and longitude coordinates. By default the tool will use a global map, but users can override this with their own GeoJSON file.~\footnote{GeoJSON is a JSON-based standard for encoding boundary and region information through GPS data. GeoJSON files for many geographic regions are easily downloadable online, and can be readily converted from shapefiles, another type of geographic boundary file.} The geography labels are analyzed in conjunction with the previously mentioned object and demographic labels, as well as external data source annotations. These can be at either the image-level, e.g., language of an image caption or region level, e.g., population size.
\end{enumerate} 


In the rest of the paper we will describe some insights automatically generated by our tool on various datasets, and potential actions that can be taken. The metrics are all run fully automatically, but based on the statistically significant results that are surfaced by the tool, we pick out the interesting findings to present in this paper that demonstrate the flavor of insight each metric will provide.


\section{Object-Based Analysis}
We begin with an object-based approach to gain a basic understanding of a dataset. Much visual recognition research has centered on recognizing objects as the core building block~\cite{Everingham10}, and a number of object recognition datasets have been collected e.g., Caltech101~\cite{caltech101}, PASCAL VOC~\cite{Everingham10}, ImageNet~\cite{imagenet, imagenet_cvpr09}. In Section~\ref{sec:obj_metrics} we introduce the metrics reported by REVISE; in Section~\ref{sec:obj_actionable} we dive into the actionable insights we surface as a result, all summarized in Table~\ref{tbl:obj_summary}.

\begin{table*}[t]
\setlength{\tabcolsep}{1.1em}
\caption{Object-based summary: for image content and object annotations of COCO}
\label{tbl:obj_summary}
\begin{center}
\begin{tabularx}{\textwidth}{@{}p{0.13\linewidth} p{0.43\linewidth} p{0.34\linewidth}@{}}\toprule \textbf{Metric} & \textbf{Example insight} & \textbf{Example action} \\
\midrule
Object counts (Sec.~\ref{sec:obj_m_counts}) & Within the supercategory \obj{appliance}, \obj{oven} and \obj{refrigerator} are overrepresented and \obj{toaster} is underrepresented & Query for more \obj{toaster} images\\
\hline
Duplicate annotations (Sec.~\ref{sec:obj_m_dup}) & The same object is frequently labeled as both \obj{doughnut} and \obj{bagel}& Manually reconcile the duplicate annotations\\
\hline
Object scale (Sec.~\ref{sec:obj_m_scale}) & \obj{Airplane} is overrepresented as very large in images, as there are few images of airplanes smaller and flying in the sky & Query more images of \obj{airplane} with \obj{kite}, since they're more likely to have a small \obj{airplane}\\
\hline
Object \mbox{co-occurrences} (Sec.~\ref{sec:obj_m_cooccur}) & \obj{Person} appears more with unhealthy \obj{food} like \obj{cake} or \obj{hot dog} than \obj{broccoli} or \obj{orange} & Query for more images of people with a healthier \obj{food}\\
\hline
Scene \mbox{diversity} (Sec.~\ref{sec:obj_m_scene})& \obj{Baseball glove} doesn't occur much outside of sports fields & Query images of \obj{baseball glove} in different scenes like a sidewalk\\
\hline
Appearance \mbox{diversity} (Sec.~\ref{sec:obj_m_appear}) & The appearance of \obj{furniture} objects become more varied when they come from scenes like \obj{water, ice, snow} and \obj{outdoor sports fields, parks} rather than predominantly from \obj{home or hotel}. & Query more images of \obj{furniture} in \obj{outdoor sports fields, parks}, since this scene is more common than \obj{water, ice, snow}, and still contributes appearance diversity\\
\bottomrule
\end{tabularx}
\end{center}
\end{table*}

\subsection{Object-based Metrics}
\label{sec:obj_metrics}

Of the metrics we will introduce, several (e.g., object counts, duplicate annotations, object scale) are commonly used by dataset collectors; others (e.g., scene or appearance diversity) are sometimes used during ad-hoc dataset examination but rarely quantified.

When the number of labels is very large (e.g., OpenImages contains 19,995) dataset analysis at the object level can be intractable to interpret. This motivates the need for higher-level \emph{supercategories}: e.g., an \obj{appliance} supercategory encompasses the more granular instances of \obj{oven}, \obj{refrigerator}, and \obj{microwave} in COCO~\cite{Lin14}. Most datasets, however, do not contain explicit mappings from labels to supercategories like COCO does. REVISE automatically bins labels to a set of predefined supercategories using the cosine similarity of word embeddings~\cite{spacy}. Results of auto-generated mappings are returned to the user, sorted by confidence, and the user is free to override any of the mappings. In a random sample of labels from the OpenImages dataset mapped to the COCO supercategories, human validation finds this automatic binning strategy to be appropriate on 44 of 50 labels.

\subsubsection{Object counts}
\label{sec:obj_m_counts}
Object counts in the real world tend to naturally follow a long-tail distribution~\cite{Ouyang16, Yang14, Salakhutdinov11}. As for object counts in datasets, there are two main views: reflecting the natural long-tail distribution (e.g., in SUN \cite{Xiao10}) or approximately equal balancing (e.g., in ImageNet~\cite{imagenet}). Either way, the first-order statistic when analyzing a dataset is to compute the per-category counts and verify that they are consistent with the target distribution. By computing how frequently an object is represented both within its supercategory, as well as among all objects, this allows for a fine-grained look at frequency statistics: for example, while the \obj{oven} and \obj{refrigerator} objects fall below the median number of instances for an object class in COCO, it is nevertheless notable that both of these objects are around twice as frequent as the average object from the \obj{appliance} class.

\begin{figure}[t]
\centering
\begin{subfigure}{.48\textwidth}
    \centering
    \includegraphics[width=.99\linewidth]{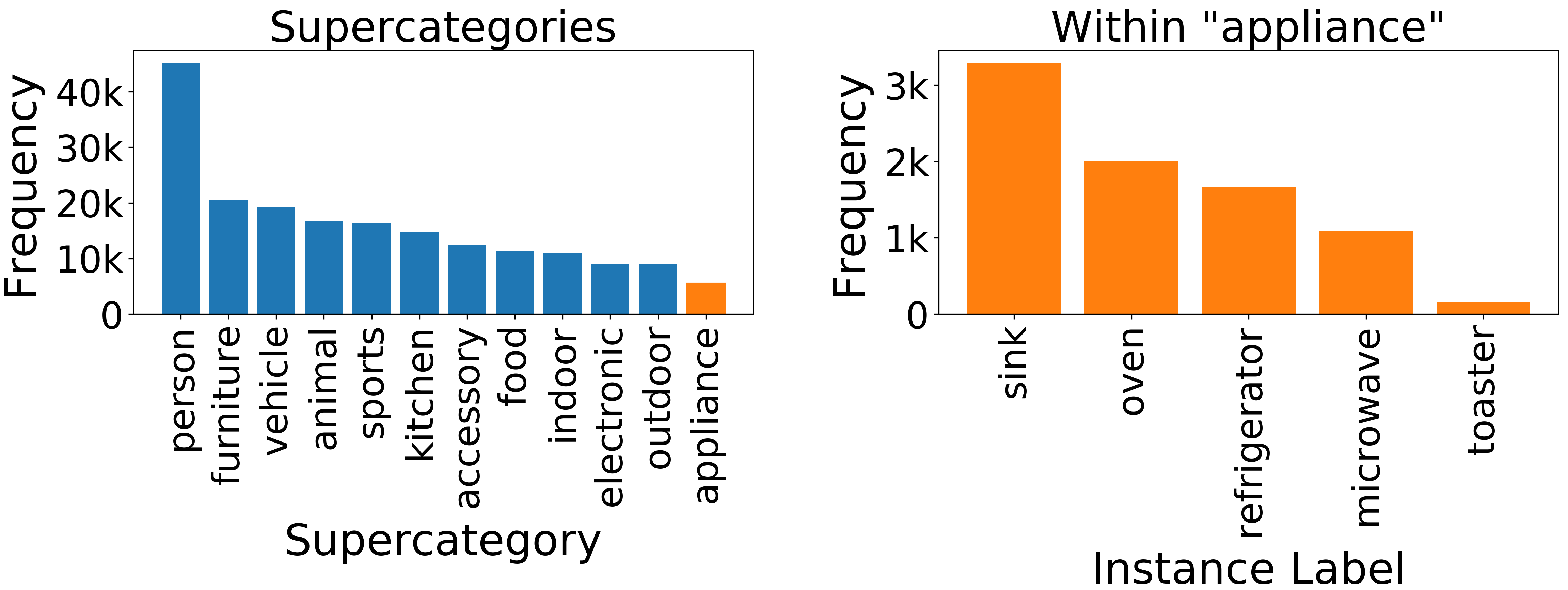}
\end{subfigure}
\caption{\obj{Oven} and \obj{refrigerator} counts fall below the median of object classes in COCO; however, they are actually over-represented within the appliance category.}
\label{fig:0_appliance}
\end{figure} 

\subsubsection{Duplicate annotations}
\label{sec:obj_m_dup}
A common issue with object dataset annotation is the labeling of the same object instance with two names (e.g., \obj{cup} and \obj{mug}), which is especially problematic in free-form annotation datasets such as Visual Genome~\cite{krishnavisualgenome}. In datasets with closed-world vocabulary, image annotation is commonly done for a single object class at a time causing confusion when the same object is labeled as both \obj{trumpet} and \obj{trombone}~\cite{imagenet}. While these occurrences are manually filtered in some datasets, automatic identification of such pairs is useful for both dataset curators (to remove errors) and to dataset users (to avoid over-counting of either object). REVISE automatically identifies such object instances. In
the OpenImages dataset \cite{openimages} some examples of automatically detected pairs include \obj{bagel} and \obj{doughnut}, \obj{jaguar} and \obj{leopard}, and \obj{orange} and \obj{grapefruit}.  In each case, the two labels are distinct (although visually similar) concepts, suggesting annotation errors.

\subsubsection{Object scale}
\label{sec:obj_m_scale}
It is well-known that object size plays a key role in object recognition accuracy~\cite{imagenet,Hoiem12}, as well as semantic importance in an image~\cite{Berg12}. While many quantizations of object scale have been proposed~\cite{Lin14, Hoiem12}, 
we aim for a metric that is both comparable across object classes and invariant to image resolution to be suitable for different datasets. Thus, for every object instance we compute the fraction of image area occupied by this instance, and quantize into 5 equal-sized bins across the entire dataset.
This binning reveals, for example, that rather than an equal 20\% for each size, 77\% of \obj{airplanes} and 73\% of \obj{pizzas} in COCO are extra large ($>9.3\%$ of the image area).

\subsubsection{Object co-occurrence}
\label{sec:obj_m_cooccur}
Object co-occurrence is a known contextual visual cue exploited by object detection models~\cite{Galleguillos08, Oliva07}, and thus can serve as an important measure of the diversity of object context. We compute all pairwise object class co-occurrence statistics within the dataset, and use them both to identify surprising co-occurrences as well as to generate potential search queries to diversify the dataset, as described in Section~\ref{sec:obj_actionable}. For example, we find that in COCO, \obj{person} appears in 43\% of images containing the \obj{food} category; however, \obj{person} appears in a smaller percentage of images containing \obj{broccoli} (15\%), \obj{carrot} (21\%), and \obj{orange} (29\%), and conversely a greater percentage of images containing \obj{cake} (55\%), \obj{donut} (55\%), and \obj{hot dog} (56\%). 
 
\subsubsection{Scene diversity}
\label{sec:obj_m_scene}
Building on quantifying the common context of an object, we additionally strive to measure the scene diversity directly. To do so, for every object class we consider the entropy of scene categories in which the object appears. 
We use a ResNet-18~\cite{He16} trained on Places~\cite{zhou2017places} to classify every image into one of 16 scene groups,\footnote{Because top-1 accuracy for even the best model on all 365 scenes is 55.19\%, but top-5 accuracy is 85.07\%, we use the less granular scene categorization at the second tier of the defined scene hierarchy \href{http://places2.csail.mit.edu/scene_hierarchy.html}{here}. For example, \obj{aquarium}, \obj{church indoor}, and \obj{music studio} fall into the scene group of \obj{indoor cultural}.} and identify objects like \obj{person} that appear in a higher diversity of scenes versus objects like \obj{baseball glove} that appear in fewer kinds of scenes (almost all baseball fields). This insight may guide dataset creators to further augment the dataset, as well as guide dataset users to want to test if their models can support out-of-context recognition on the objects that appear in fewer kinds of scenes, for example baseball gloves in a street.


\subsubsection{Appearance diversity}
\label{sec:obj_m_appear}
Finally, we consider the appearance diversity (i.e., intra-class variation) of each object class, which is a primary challenge in object detection~\cite{Yao17}. We use a ResNet-110 network~\cite{resnetgithub} trained on CIFAR-10~\cite{cifar10} to extract a 64-dimensional feature representation of every instance bounding box, resized to 32x32 pixels. We first validate that distances in this feature space correspond to semantically meaningful measures of diversity. To do so, on the COCO dataset we compute the average distance with $n=500,000$ between two object instances of the same class ($5.91\pm1.44$), and verify that it is smaller than the average distance between two object instances belonging to different classes but the same supercategory ($6.24\pm1.42$), with a Cohen's D effect size of $.23$ and further smaller than the average distance between two unrelated objects ($6.48\pm 1.44$), with a Cohen's D effect size of $.17$. This metric allows us to identify individual object instances that contribute the most to the diversity of an object class, and informs our interventions in the next section.

\subsection{Object-based Actionable Insights}
\label{sec:obj_actionable}

The metrics of Section~\ref{sec:obj_metrics} help surface biases or other issues, but it may not always be clear how to address them. We strive to mitigate this concern by providing examples of meaningful, actionable, and useful steps to guide the user.

For duplicate annotations, the remedy is straight-forward: perform manual cleanup of the data, e.g., as in Appendix E of~\cite{imagenet}. For the others the path is less straight-forward. For datasets that come from web queries, following the literature~\cite{Everingham10, imagenet, Lin14} REVISE defines search queries of the form ``\texttt{XX} and \texttt{YY},'' where \texttt{XX} corresponds to the target object class, and \texttt{YY} corresponds to a contextual term (another object class, scene category, etc.). REVISE ranks all possible queries to identify the ones that are most likely to lead to the target outcome, and we investigate this approach more thoroughly in
Appendix C.

\begin{figure}[t]
\centering
\begin{subfigure}{.26\textwidth}
    \centering
    \includegraphics[height=3.5cm]{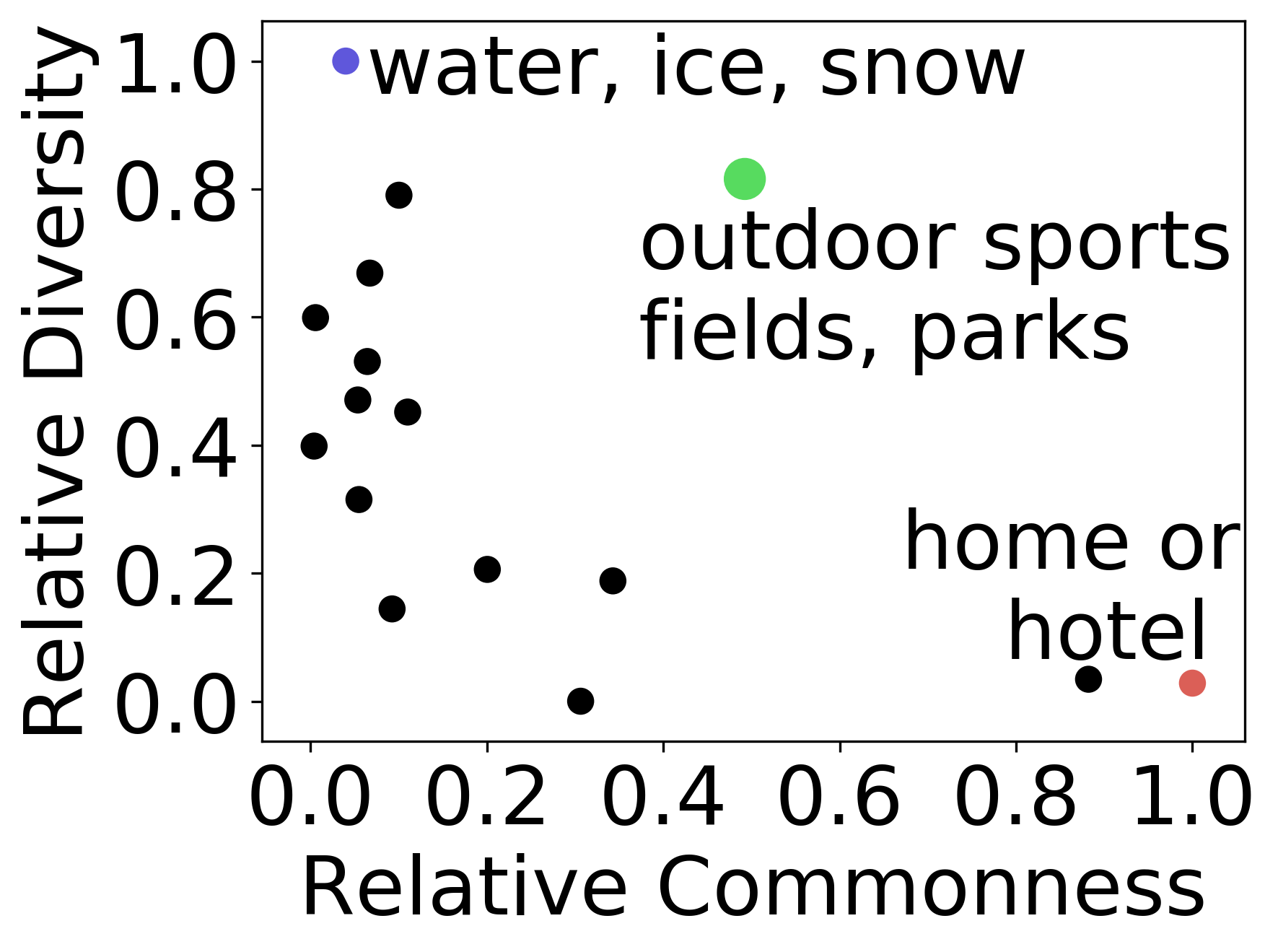}
\end{subfigure}
\begin{subfigure}{.72\textwidth}
    \includegraphics[height=2.7cm]{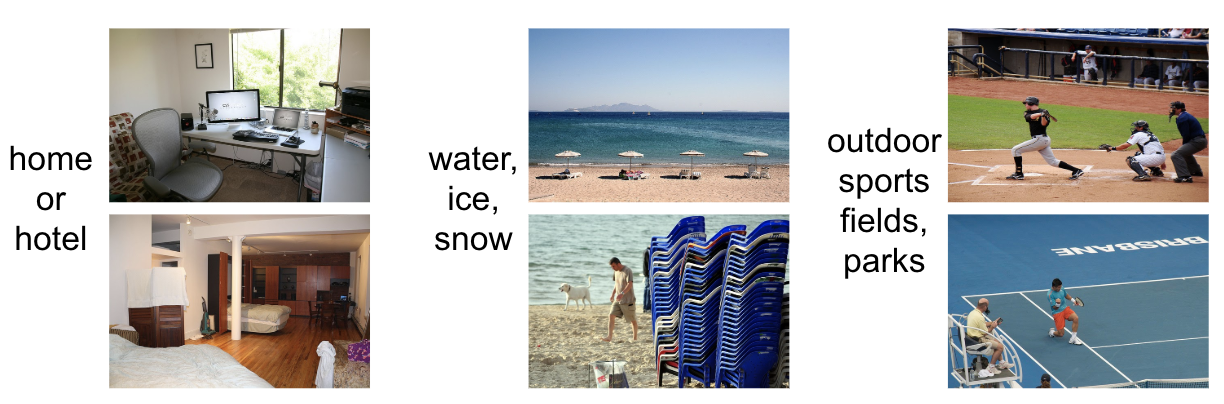}
\end{subfigure}
\caption{The top shows the tradeoff for \obj{furniture} in COCO between how much scenes increase appearance diversity (our goal) and how common they are (ease of collecting this data). To maximize both, \obj{outdoor sports fields, parks} would be the most efficient way of augmenting this category. \obj{Water, ice, snow} provides the most diversity but is hard to find, and \obj{home or hotel} is the easiest to find but provides little diversity. On the bottom are sample images of \obj{furniture} from these scenes.}
\label{fig:9_co_scatter}
\end{figure}

For example, within COCO, \obj{airplanes} have low diversity of scale and are predominantly large in the images. Our tool identifies that smaller airplanes co-occurred with objects like \obj{surfboard} and scenes like \obj{mountains, desert, sky} (which are more likely to be photographed from afar). In other words, size matters by itself, but a skewed size distribution could also be a proxy for other types of biases.
Dataset creators aiming to diversify their dataset towards a more uniform distribution of object scale can use these queries as a guide. These pairwise queries can similarly be used to diversify appearance diversity. \obj{Furniture} objects appear predominantly in indoor scenes like \obj{home or hotel}, so querying for furniture in scenes like \obj{water, ice, snow} would diversify the dataset. However, this combination is quite rare, so we want to navigate the tradeoff between a pair's commonness and its contribution to diversity. Thus, we are more likely to be successful if we query for images in the more common \obj{outdoor sports fields, parks} scenes, which also brings a significant amount of appearance diversity. The tool provides a visualization of this tradeoff (Fig.~\ref{fig:9_co_scatter}), allowing the user to make the final decision.

\section{Person-Based Analysis}

We next look into discrepancies in various aspects of how people of differing demographic attributes are represented, summarized in Table~\ref{tbl:gen_summary}. The datasets we consider here are COCO~\cite{Lin14}, for which we have gender and skin tone annotations, and OpenImages~\cite{openimages}, for which we have gender annotations. In Section~\ref{sec:gen_metrics} we explain some of the metrics that we collect, and in Section~\ref{sec:gen_actionable} we discuss possible actions.

\smallsec{Gender labels} The gender labels in COCO are from~\textcite{Zhao17}, and their methodology in determining the gender for an image is that if at least one caption contains the word ``man" and there is no mention of ``woman", then it is a male image, and vice versa for female images. We use the same methodology along with other gendered labels like ``boy" and ``girl" on OpenImages' pre-existing annotations of individuals. It is important to acknowledge that the labels we are using are those of perceived binary gender, which is not  inclusive of all gender categories. We will use the terms male and female to refer to binarized socially-perceived gender expression, and not gender identity nor sex assigned at birth, neither of which can be inferred from an image. In Appendix~\ref{app:per_m_infer} we consider some of the problems that arise from using gender labels that have been inferred in this way.

\smallsec{Skin tone labels} Our skin tone annotations for COCO come from~\cite{zhao2021captionbias}, and are numbered 1-6 according to the Fitzpatrick scale \cite{fitzpatrick}, where 1 is the lightest and 6 is the darkest. We use perceived skin tone as a poor proxy for race, and acknowledge that this risks reifying a particular inaccurate conception of race~\cite{hanna20race}. We consider skin tone as an ordinal variable, and analyze trendlines that result as we increase or decrease along this axis.

\begin{table*}[t]
\ra{1.2}
\setlength{\tabcolsep}{1.1em}
\caption{Person-based summary: investigating representation of people with different demographic attributes.}
\label{tbl:gen_summary}
\begin{center}
\begin{tabularx}{\textwidth}{@{}p{0.13\linewidth} p{0.43\linewidth} p{0.34\linewidth}@{}}\toprule \textbf{Metric} & \textbf{Example insight} & \textbf{Example action} \\
\midrule
Person Prominence (Sec.~\ref{sec:per_m_prom}) & As the skin tone of the person in an image increases in darkness, the person is more likely to be smaller and further from the center. & Collect more images of people of different skintones as the subject of an image rather than in the background.\\
\hline
Contextual representation (Sec.~\ref{sec:per_m_context}) & Males occur in more outdoors scenes and with \obj{sports} objects. Females occur in more indoors scenes and with \obj{kitchen} objects. & Collect more images of females in outdoors scenes with \obj{sports} objects, and vice versa for males.\\
\hline
Instance counts \mbox{and distances} (Sec.~\ref{sec:per_m_countdist}) & In images with musical instrument \obj{organ}, males are more likely to be actually playing the \obj{organ}. & Collect more images of females playing \obj{organs}.\\
\hline
Appearance differences (Sec.~\ref{sec:per_m_appear})& Males in \obj{sports uniforms} tend to be playing outdoor sports, while females in \obj{sports uniforms} are often indoors or in swimsuits. & Collect more images of each gender with \obj{sports uniform} in their underrepresented scenes.
\\
\bottomrule
\end{tabularx}
\end{center}
\end{table*}

\subsection{Person-based Metrics}
\label{sec:gen_metrics}
In this section, we will give representative findings for each metric that demonstrate the kind of insight our tool can provide. We start out by considering both gender and skin tone for COCO in Sec.~\ref{sec:per_m_prom} and \ref{sec:per_m_context}, before transitioning to gender in OpenImages in Sec.~\ref{sec:per_m_countdist} and \ref{sec:per_m_appear}.

\subsubsection{Person Prominence}
\label{sec:per_m_prom}
\begin{figure}
\centering
\includegraphics[width=0.48\textwidth]{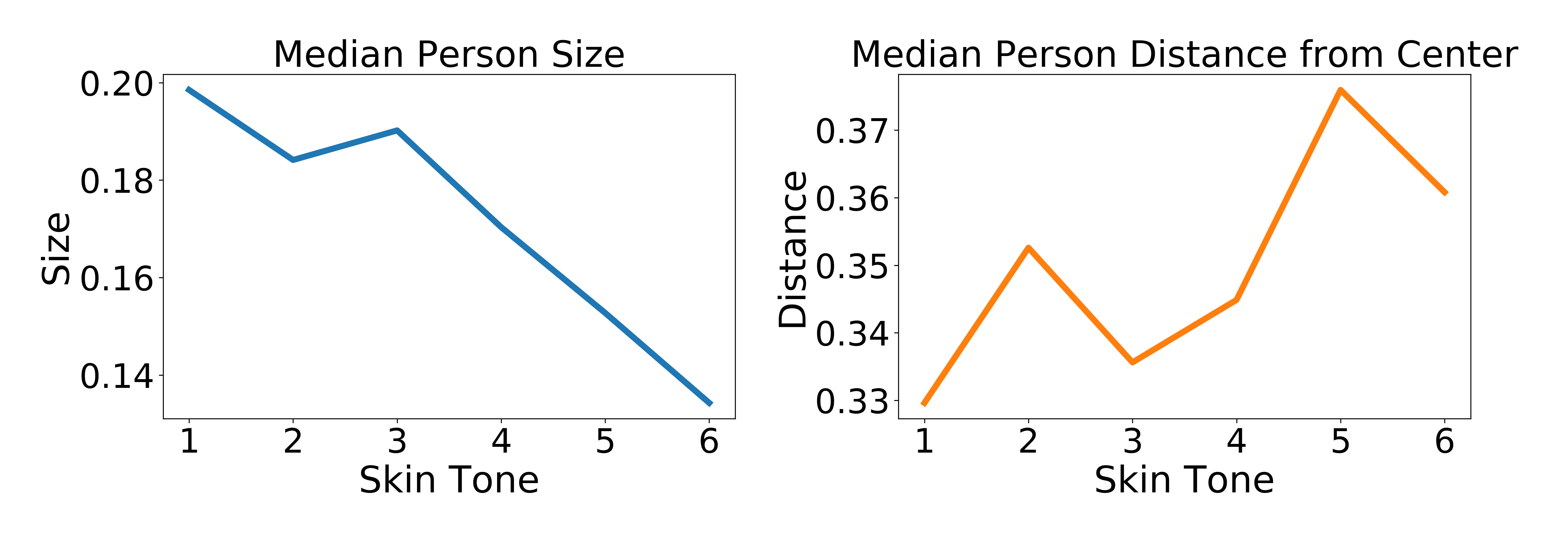}
\caption{In the COCO dataset, as a person's skin tone increases in darkness, that person is more likely to be smaller and further from the center. This indicates that people of darker skin tones are more likely to be in the background of an image rather than featured prominently. We used Jonckheere's trend test~\cite{jonckheere1954test} to show there is an a priori ordering to size and distance values by skin tone with p-values of $2.11\mathrm{e-}7$ and $.014$, respectively.}
\label{fig:coco_siz}
\end{figure}

As our first line of analysis regarding how people of different demographic attributes are represented, we consider the proportion of an image a person takes up, as well as their distance from the center. We treat these two measures as a proxy for importance~\cite{Berg12}, where people who are larger and more to the center of an image are the focal point. We run the analysis for COCO on people differentiated both by gender and by skin tone. For gender, people who are male tend to take up more of the image ($.268\pm.213$ for male vs $.138\pm.148$ for female, with a Cohen's D effect size of .709) and be closer to the center ($.363\pm.218$ for male vs $.510\pm.250$ for female, with a Cohen's D effect size of .627). For people of different skin tones, in Fig.~\ref{fig:coco_siz} we see that as skin tone increases in darkness, the person is more likely to take up less of the image, as well as be further from the center. This indicates a bias against females and people of darker skin tones as being less likely to be the focal point of an image.

\subsubsection{Contextual representation}
\label{sec:per_m_context}
\begin{figure*}[t]
\centering
\begin{subfigure}{.52\textwidth}
    \centering
    \includegraphics[height=7.7cm, clip]{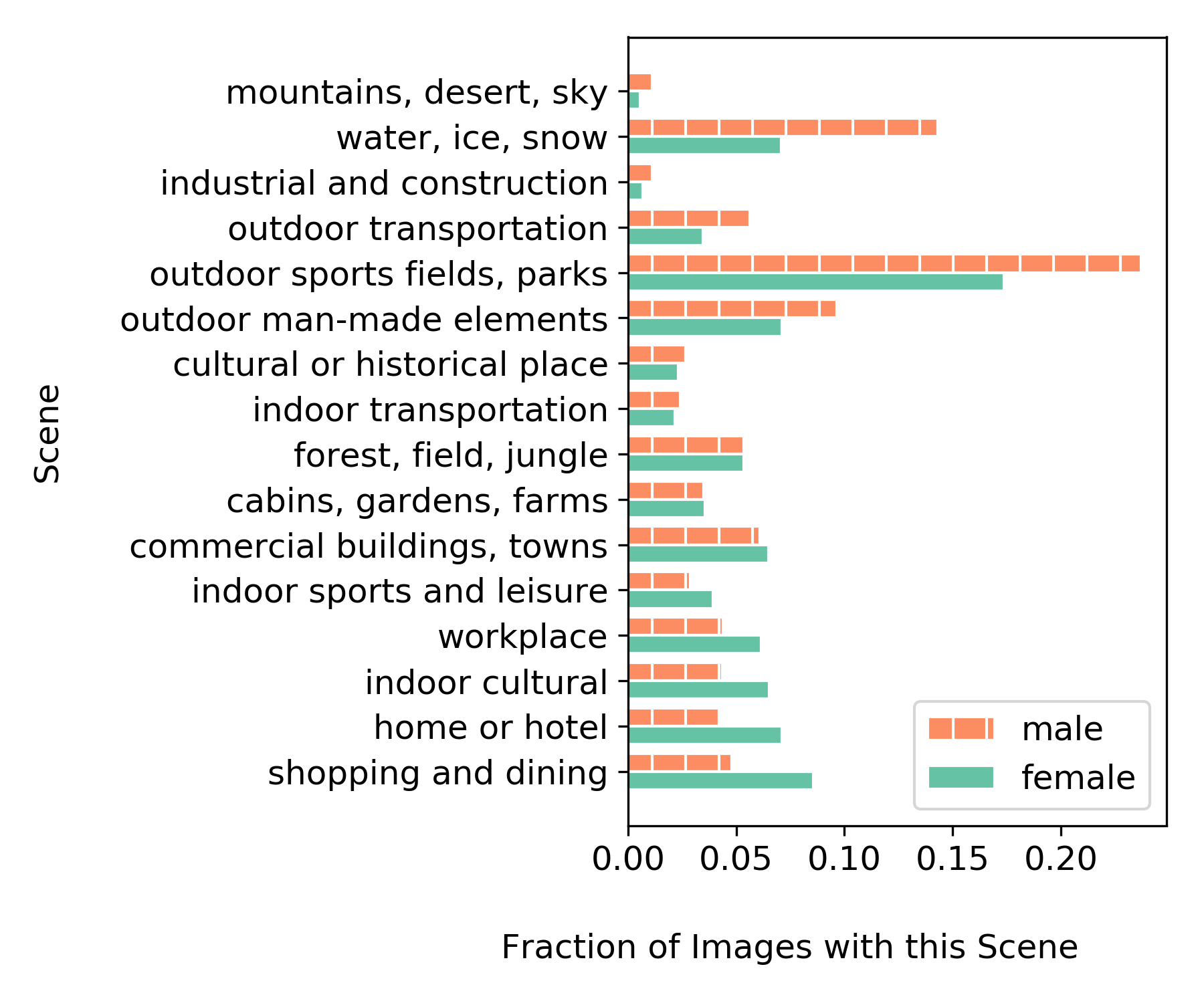}
\end{subfigure}
\begin{subfigure}{.44\textwidth}
    \centering
    \includegraphics[height=7.7cm, clip]{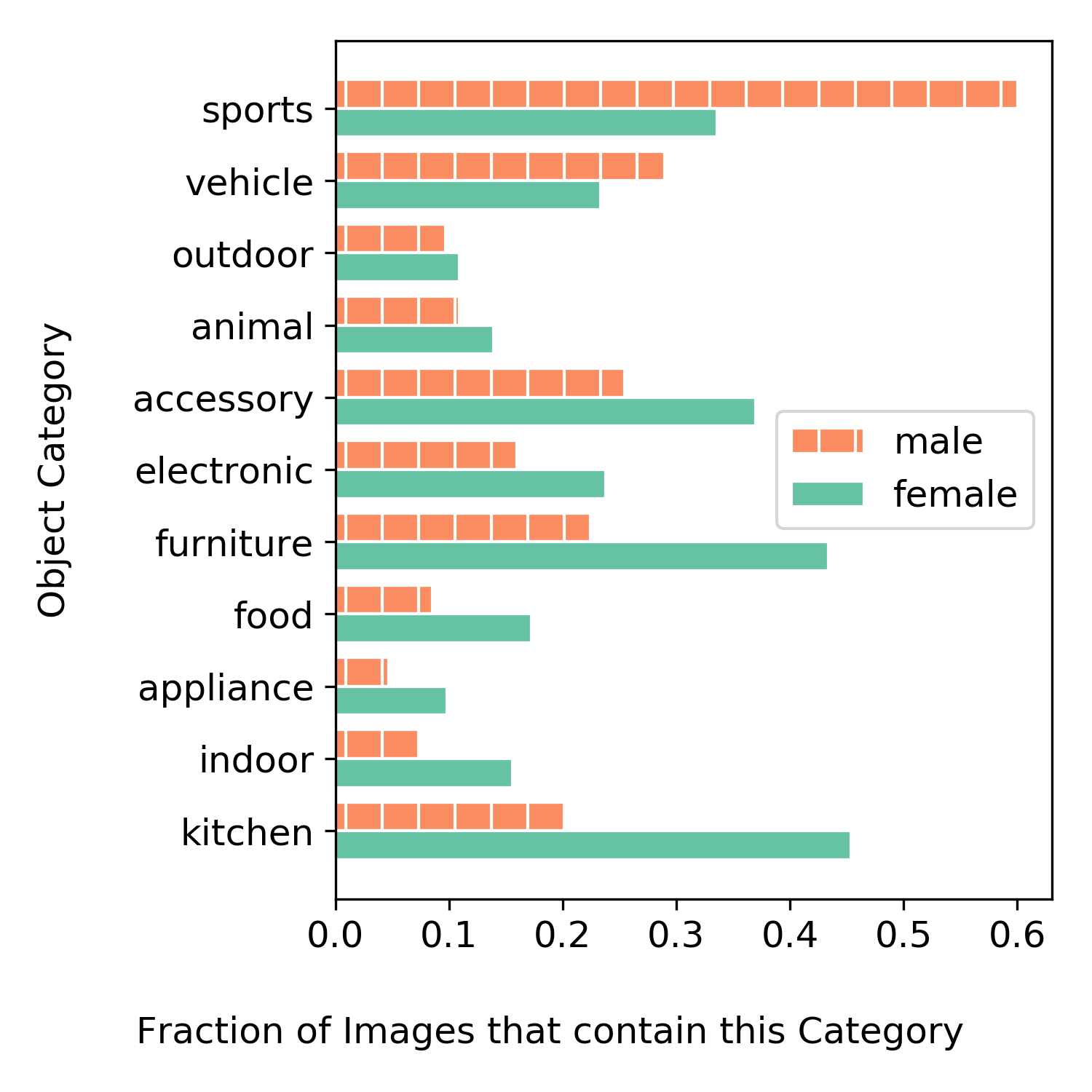}
\end{subfigure}
\caption{Contextual information of images in COCO by gender, represented by fraction that are in a scene (left) and have an object from the category (right).}
\label{fig:co_context}
\end{figure*}

Looking beyond just the person themselves, we consider the contexts that people with different demographic attributes tend to be featured in through the object groups they cooccur with, and the scenes they appear in. We first consider people of two different genders in COCO, and in Fig.~\ref{fig:co_context} see that images with females tend to be more indoors in scenes like \obj{shopping and dining} and with object groups like \obj{furniture}, \obj{accessory}, and \obj{appliance}. On the other hand, males tend to be in more outdoors scenes like \obj{sports fields} and \obj{water, ice, snow}, and with object groups like \obj{sports} and \obj{vehicle}. These trends reflect gender stereotypes in many societies and can propagate into the models. While there is work on algorithmically intervening to break these associations, there are often too many proxy features to robustly do so. Thus it can be useful to intervene at the dataset creation stage.

\begin{figure}
\centering
\includegraphics[width=0.48\textwidth]{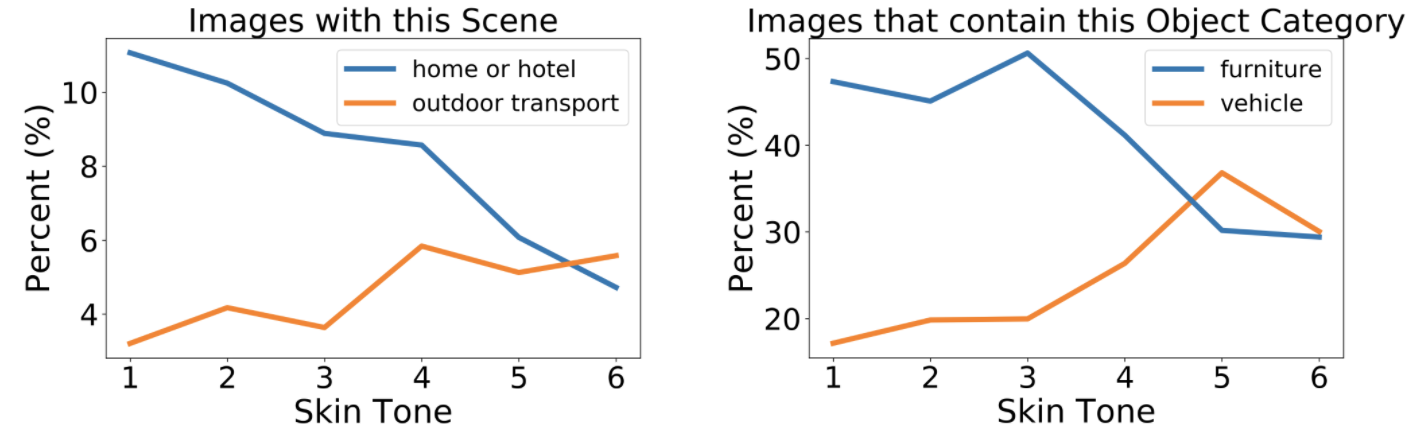}
\caption{We fit regression lines between co-occurrences of people with particular skin tones, and scenes and object categories. We show in the figure example categories where the Wald test has $p<.05$ that the slopes are non-zero, revealing trends that appear in image context as skin tone changes. On the left, we see that as an individual's skin tone increases in darkness, they are less likely to be pictured in \obj{home or hotel} scenes, and more likely to be pictured in \obj{outdoor transportation} scenes. On the right, we see that for object categories, people with darker skin tones are less likely to be pictured with \obj{furniture} objects, and more likely to be pictured with \obj{vehicle} objects.}
\label{fig:coco_skin}
\end{figure}

Then, we consider these analyses in COCO along the ordinal variable of skin tone. In Fig.~\ref{fig:coco_skin} we see statistically significant trends according to the Wald test on a non-zero slope of regression lines where people with lighter skin tones are more likely to be in \obj{home or hotel} scenes and with object groups like \obj{furniture}, and people with darker skin tones are more likely to be in \obj{outdoor transportation} scenes and with object groups like \obj{vehicle}. In the next metric we dig deeper into these object categories by considering the particular objects themselves.

\subsubsection{Instance Counts and Distances} 
\label{sec:per_m_countdist}
Analyzing object instances allows a more granular understanding of biases in the dataset. For example, in the previous metric on COCO we found \obj{vehicle} objects to occur more with people of darker skin tones, and \obj{furniture} with people of lighter skin tones. The specific \obj{vehicle} objects that fit this trend are \obj{motorcycle} and \obj{bus}, while the specific \obj{furniture} objects are \obj{bed} and \obj{couch}. 

\begin{figure}
\centering
\includegraphics[width=0.49\textwidth]{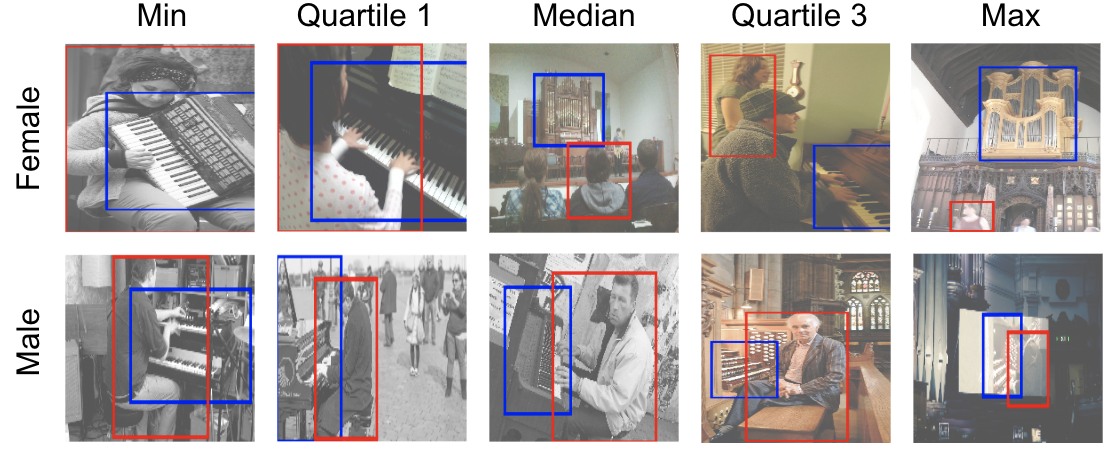}
\caption{5 images from OpenImages for a person (red bounding box) of each gender pictured with an organ (blue bounding box) along the gradient of inferred 3D distances. Males tend to be featured as actually playing the instrument, whereas females are oftentimes merely in the same space as the instrument.}
\label{fig:3_oi_qual}
\end{figure}

In OpenImages we find that objects like \obj{cosmetics}, \obj{doll}, and \obj{washing machine} are overrepresented with females, and objects like \obj{rugby ball}, \obj{beer}, \obj{bicycle} are overrepresented with males. However, beyond just looking at the number of times objects appear, we also look at the distance an object is from a person. We use a scaled distance measure as a proxy for understanding if a particular person, $p$, and object, $o$, are actually interacting with each other in order to derive more meaningful insight than just quantifying a mutual appearance in the same image. The distance measure we define is 
\begin{equation}
    dist = \frac{\textrm{distance between p and o centers}}{\sqrt{\textrm{area}_{\mathrm{p}}*\textrm{area}_{\mathrm{o}}}}
\end{equation}
to estimate distance in the 3D world, where $\textrm{area}_{\textrm{p}}$ is measured on a normalized image of total area 1.
In Appendix B 
we validate this notion that our distance measure can be used as a proxy interaction. We consider these distances in order to disambiguate between situations where a person is merely in an image with an object in the background, rather than directly interacting with the object, revealing biases that were not clear from just looking at the frequency differences. For example, \obj{organ} (the musical instrument) did not have a statistically significant difference in frequency between the genders, but does in distance, or under our interpretation, relation. In Fig.~\ref{fig:3_oi_qual} we investigate what accounts for this difference and see that when a male person is pictured with an organ, he is likely to be playing it, whereas a female person may just be near it but not necessarily directly interacting with it. Through this analysis we discover something more subtle about how an object is represented.

\subsubsection{Appearance Differences}
\label{sec:per_m_appear}
\begin{figure}[t]
\centering
\includegraphics[height=3.6cm]{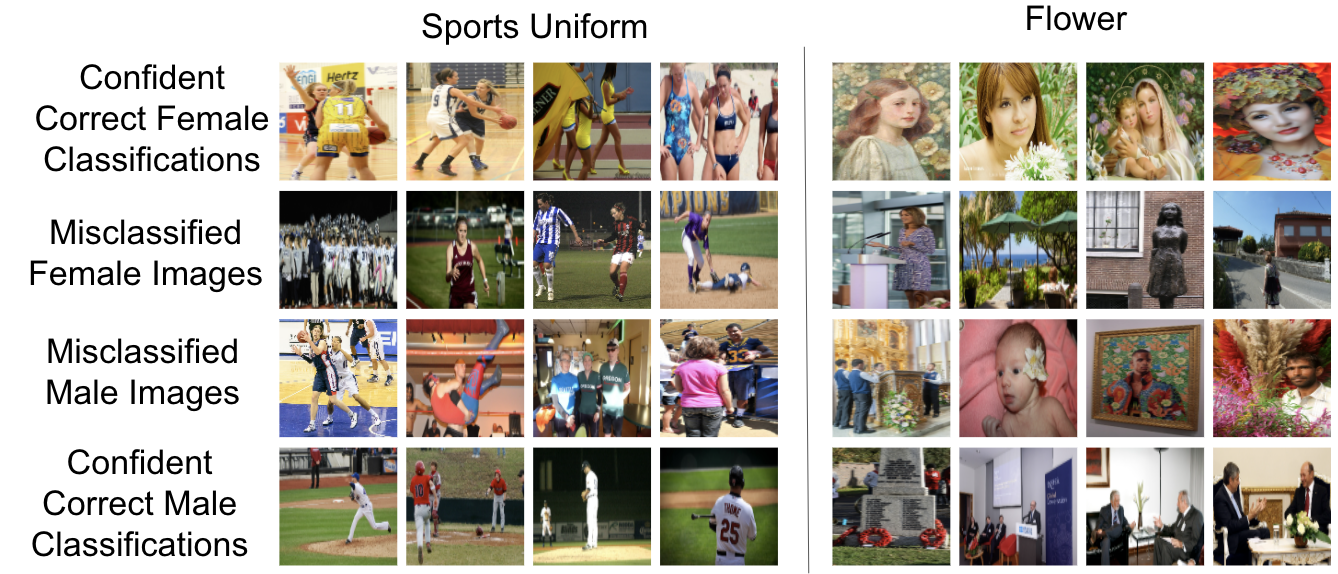}
\caption{Qualitative interpretation of what the visual model has learned for the \obj{sports uniform} and \obj{flower} objects between the two genders in OpenImages. ``Confident Correct" are the images with the highest confidence scores.}
\label{fig:4_co_svm}
\end{figure}

We also look into the appearance differences in images of each gender with a particular object. This is to further disambiguate situations where occurrence counts, or even distances, aren't telling the whole story. This analysis is done by (1) extracting FC7 features from AlexNet~\cite{alexnet} pretrained on Places \cite{zhou2017places} on a randomly sampled subset of the images to get scene-level features, (2) projecting them into $\sqrt{\textrm{number of samples}}$ dimensions (as is recommended in~\cite{Hua04, Jain78}) to prevent over-fitting, and then (3) fitting a Linear Support Vector Machine to see if it is able to learn a difference between images of the same object with different genders. To make sure the female and male images are actually linearly separable and the classifier isn't over-fitting, we look at both the accuracy as well as the ratio in accuracy between the SVM trained on the correctly labeled data and randomly shuffled data. In Fig.~\ref{fig:4_co_svm} we can see what the Linear SVM has learned on OpenImages for the \obj{sports uniform} and \obj{flower} categories. For \obj{sports uniform}, males tend to be represented as playing outdoor sports like baseball, while females tend to be portrayed as playing an indoor sport like basketball or in a swimsuit. For \obj{flower}, we see another drastic difference in how males and females are portrayed, where males pictured with a \obj{flower} are in formal, official settings, whereas females are in staged settings or paintings.


\subsection{Person-based Actionable Insights}
\label{sec:gen_actionable}

Compared to object-based metrics, the actionable insights for person-based metrics are less concrete and more nuanced. There is a tradeoff between attempting to represent the visual world as it is versus as we think it should be. For example, in contemporary societies, gender representation in various occupations, activities, etc. is unequal, so it is not obvious that aiming for gender parity across all object categories is the right approach. Biases that are systemic and historical are more problematic than others~\cite{Bearman09}, and this analysis cannot be automated. Further, the downstream impact of unequal representation depends on the specific models and tasks.
Nevertheless, we provide some recommendations.

A trend that appeared in the metrics is that images frequently fell in line with common gender and racial stereotypes. Each group of people was under- or over-represented in a particular way, and dataset collectors may want to adjust their datasets to account for these by augmenting in the direction of the underrepresentations. Dataset users may want to audit their models, and look into to what extent their models have learned the dataset's biases before they are deployed.

\section{Geography-Based Analysis}
Finally, we turn to the geography of the images. We consider geography in the context of the object-based and person-based analyses from before, as well as additional axes. Geography uniquely interacts with both the types of objects that appear in images, as well as the demographics of the people. Because of these interactions, biases and problems around generalization have been shown to appear~\cite{Shankar17, gebru2017street, Devries19}. 

In addition to COCO, for which we can derive geography labels on a subset of the images by querying the source of the images, i.e., Flickr,  we also consider the global YFCC100m dataset\footnote{We use different subsets of the YFCC100m dataset depending on the particular annotations required by each metric.}~\cite{Thomee16}, and the New York-centric BDD100K~\cite{yu2020bdd} self-driving car dataset.\footnote{We consider the subset of the BDD100K dataset with images in New York City, which is  a majority of the dataset.} 

In Sec.~\ref{sec:geo_metrics} we present findings from our metrics, and in Sec.~\ref{sec:geo_actionable} we discuss what can be done about them.

\begin{table*}[t]
\ra{1.2}
\setlength{\tabcolsep}{1.1em}
\caption{Geography-based summary: looking into the geo-representation of a dataset, and how that differs between different regions}.
\label{tbl:geo_summary}
\begin{center}
\begin{tabularx}{\textwidth}{@{}p{0.13\linewidth} p{0.43\linewidth} p{0.34\linewidth}@{}}\toprule \textbf{Metric} & \textbf{Example insight} & \textbf{Example action} \\
\midrule
Geography \mbox{distribution} (Sec.~\ref{sec:geo_m_dist}) & Most images are from the USA, with very few from the countries of Africa & Collect more images from the countries of Africa\\
\hline
Geography \mbox{by object} (Sec.~\ref{sec:geo_m_obj}) & \obj{Wildlife} is overrepresented in Kiribati, and \obj{mosque} in Iran & Compare dataset frequencies to real-world frequencies; consider collecting other kinds of images representing these countries\\
\hline
Geography \mbox{by people} (Sec.~\ref{sec:geo_m_people}) & Underrepresented regions like Africa and South Asia contain many of the images of people with darker skin tones & Collect more images from underrepresented regions to also diversify the people of different skin tones being represented.\\
\hline
Geography \mbox{by language} (Sec.~\ref{sec:geo_m_lang}) & Countries in Africa and Asia that are already underrepresented are frequently represented by non-locals rather than locals & Collect more images taken by locals rather than visitors in underrepresented countries\\
\hline
Geography \mbox{by income} (Sec.~\ref{sec:geo_m_income}) & Normalized by square mile, wealthier zip codes have more images, which also contain a different distribution of labels & Collect more images from zip codes with lower incomes\\
\hline
Geography \mbox{by weather} (Sec.~\ref{sec:geo_m_weather}) & Northern California has significantly less \obj{snowy} images than New York City & Finetune a model on a weather distribution most similar to that in which it will be deployed\\
\bottomrule
\end{tabularx}
\end{center}
\end{table*}

\subsection{Geography-based Metrics}
\label{sec:geo_metrics}
In this section we analyze geography in the context of objects and people appearances, but also language, income, and weather. For distribution (Sec.~\ref{sec:geo_m_dist}), objects (Sec.~\ref{sec:geo_m_obj}), and language (Sec.~\ref{sec:geo_m_lang}) we look at the YFCC100m dataset, for people (Sec.~\ref{sec:geo_m_people}) we look at COCO, and then for income (Sec.~\ref{sec:geo_m_income}) and weather (Sec.~\ref{sec:geo_m_weather}) we look at BDD100K. Additionally, our analysis on geography by income is a case study into what our automated analyses in conjunction with an external data source of region-level labels may look like. One could also imagine plugging in a different external data source, e.g., region-level population size, and the tool would automatically run the same metrics along this axis instead.

\subsubsection{Geographic distribution}
\label{sec:geo_m_dist}
\begin{figure}[t]
\centering
\begin{subfigure}{.48\textwidth}
    \centering
    \includegraphics[width=.92\linewidth]{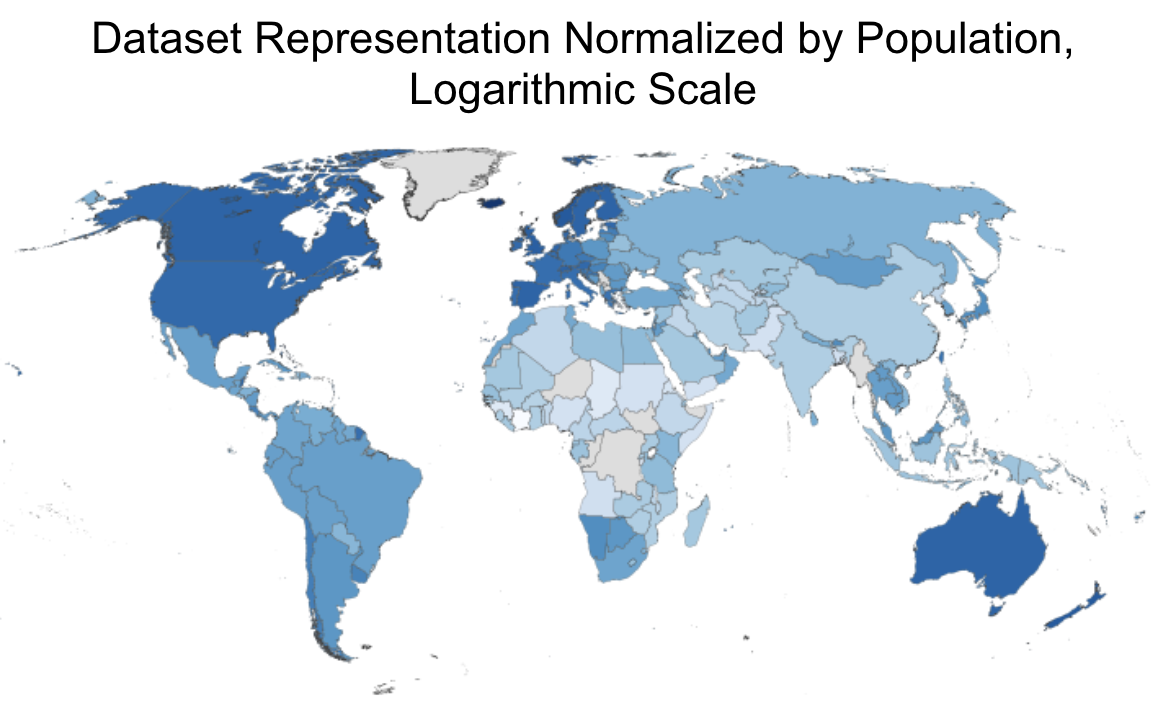}
\end{subfigure}
\caption{Geographic distribution normalized by population in YFCC100m.}
\label{fig:yf_maps}
\end{figure}

%
The first line of analysis is to look at the overall geographic distribution of a dataset. Researchers have looked at OpenImages and ImageNet and found these datasets to be Amerocentric and Eurocentric~\cite{Shankar17}, with models dropping in performance when being run on images from underrepresented locales. In Fig.~\ref{fig:yf_maps} it immediately stands out that in the global YFCC100m dataset, the USA is drastically overrepresented compared to most other countries, with the continent of Africa being very sparsely represented. This can lead to generalization problems where a model may perform worse on image from a region it has not seen as much of~\cite{Devries19}.

\subsubsection{Geography by Object}
\label{sec:geo_m_obj}
In the YFCC100m dataset, we have access to image tags, which we treat as object labels. We combine our object-based analysis techniques with this geography data, allowing us to discern if certain labels are over- or under-represented between different areas. We then begin by considering the frequency with which each image tag appears in the set of a country's tags, compared to the frequency that same tag makes up in the rest of the countries. Some examples of over- and under- representations include Kiribati with \obj{wildlife} at 86x, Iran with \obj{mosque} at 30x, Egypt with \obj{politics} at 20x, and United States with \obj{safari} at .92x. We note that, as seen in the previous metric, this dataset is so skewed in terms of representation that most statistically significant underrepresentations are in the United States, as no other country has a high enough sample size. Additionally, whether these over- or under-representations are problematic enough to warrant intervention is entirely up to the user and their downstream task. We have normalized these tags by number of tag occurrences, and \emph{not} by real-world distributions of the objects they mention, e.g., perhaps there are simply more \obj{mosques} in Iran than other countries and this overrepresentation is innocuous and in fact a representative depiction of the country --- it is up to the user to verify this.

\begin{figure}[t]
\centering
\includegraphics[height=.96cm]{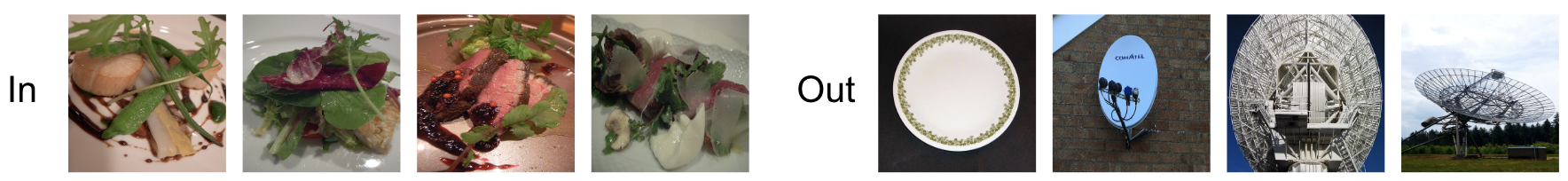}
\caption{A qualitative look at YFCC100m for what the visual model confidently and correctly classifies for images with the \obj{dish} tag as in Eastern Asia, and out.}
\label{fig:6_yf_qual}
\end{figure}

We also look beyond the numbers themselves into the appearances of how different subregions, as defined by the United Nations geoscheme~\cite{ungeoscheme}, represent certain tags. \textcite{Devries19} showed that object-recognition systems perform worse on images from countries that are not as well-represented in the dataset due to appearance differences within an object class, so we look into such appearance differences within a Flickr tag. We perform the same analysis as in Sec.~\ref{sec:gen_metrics} where we run a Linear SVM on the featurized images, this time performing 17-way classification between the different subregions. In Fig.~\ref{fig:6_yf_qual} we show an example of the \obj{dish} tag, and what images from the most accurately classified subregion, Eastern Asia, look like compared to images from the other subregions. Images with the \obj{dish} tag tend to refer to food items in Eastern Asia, rather than a satellite dish or plate, which is a more common practice in other regions. This example is telling of a more pernicious problem than mis-identifying dishes, which is that of dialect differences between regions, and how that might affect the semantic meaning of a label. Disentangling homonyms will require computer vision systems to pay attention to the more subtle nuances of linguistics~\cite{roll2018homonyms}. It may be important to know if
tags are represented differently across subregions so that models do not overfit to one particular subregion's representation of an object.

\subsubsection{Geography by People}
\label{sec:geo_m_people}
\begin{figure}[t]
\centering
\includegraphics[height=4cm]{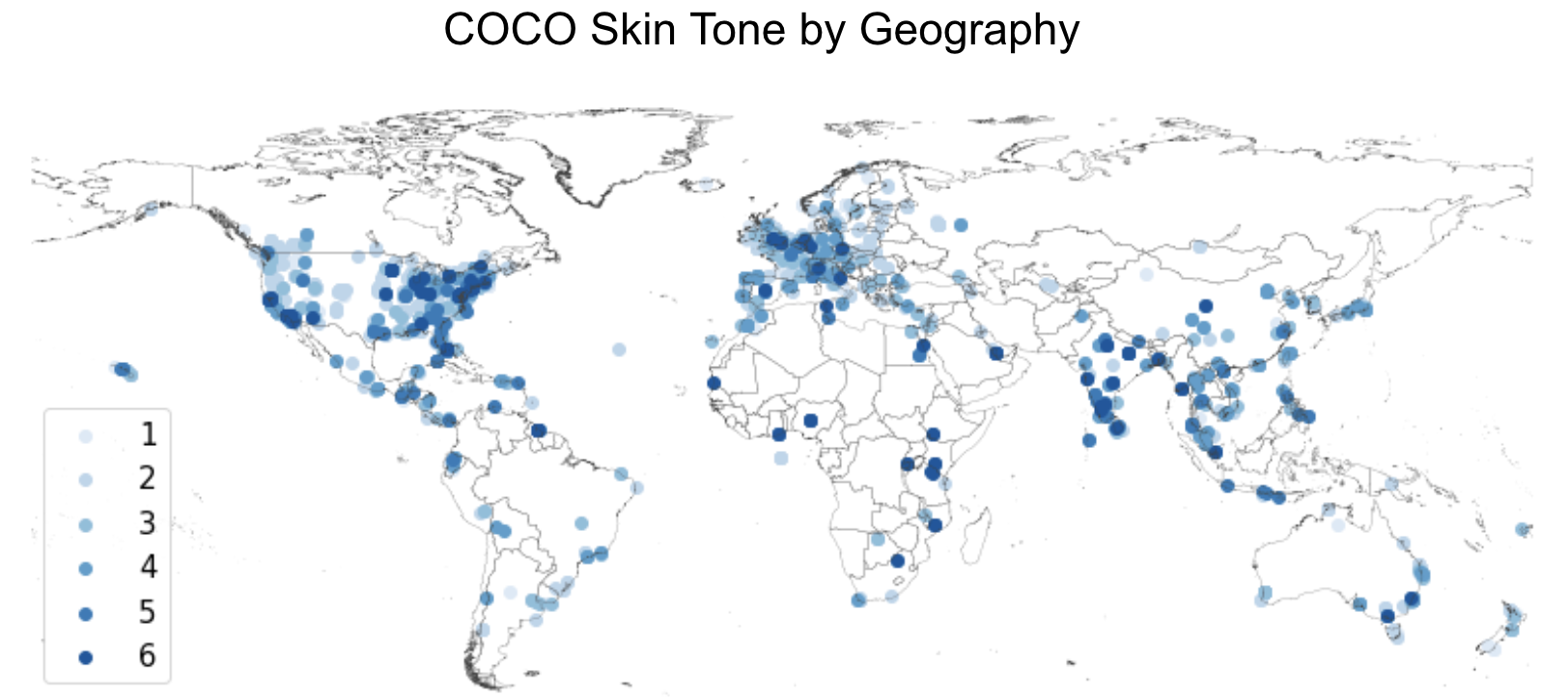}
\caption{Geographical distribution of COCO images based on skin tone annotations. Images from South Asia and Africa tend to contain people of darker skin tones, although the majority of images are coming from the United States and Europe.}
\label{fig:coco_attgeo}
\end{figure}

Next, we combine our COCO demographic skin tone annotations with geography labels. In Fig.~\ref{fig:coco_attgeo} we see that images of people with darker skin tones tend to come from South Asia and Africa, but neither of these regions are very well-represented compared to images from the United States and Europe. In fact, while 85.5\% of images of people with lighter skin tones (values 1-3) come from North America and Europe, this number is 58.2\% for people with darker skin tones (values 4-6). Models that use this dataset may develop an understanding of people with darker skin tones that will be primarily informed by people from North America and Europe, which is a very small sample of people with darker skin tones in the world. Cultural practices differ among people across regions, and depending on the downstream application, it could be important that an understanding of a group not be informed only by the people in one geographic region.

For this particular dataset, we can additionally customize our tool to incorporate an external data source. Looking at the images only within the United States and binning by urban centers, as defined by the U.S. Census, we find that while 84.4\% of images of people with lighter skin tones 1-3 are located in an urban area, 92.7\% of images of people with darker skin tones 4-6 are located in an urban area.

\subsubsection{Geography by Language}
\label{sec:geo_m_lang}
When we looked at the global distribution of the 

\begin{figure}[t]
\centering
\begin{subfigure}{.48\textwidth}
    \centering
    \includegraphics[width=.96\linewidth]{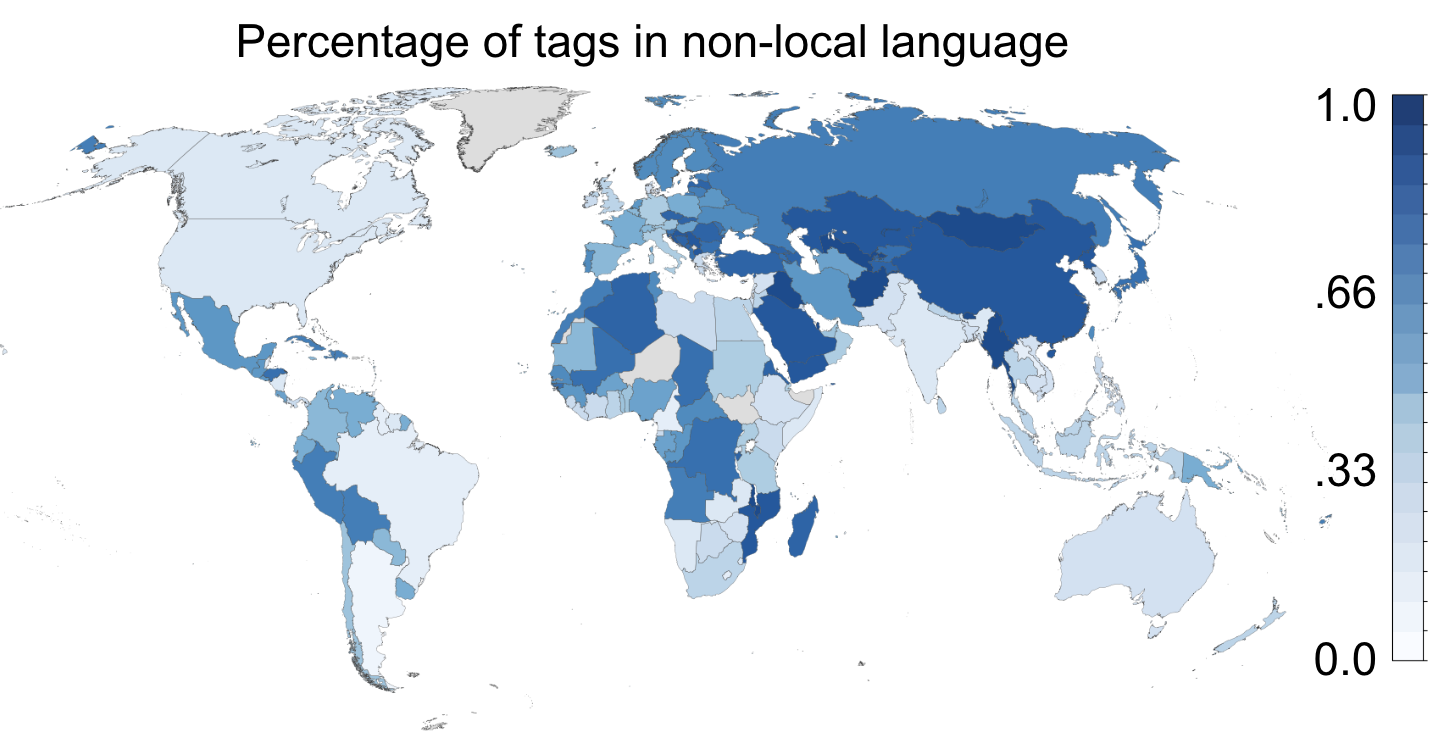}
\end{subfigure}
\caption{Percentage of tags in a non-local language in YFCC100m. Even when underrepresented countries are imaged, it is not necessarily by someone local to that area.}
\label{fig:yf_lang}
\end{figure}

\noindent YFCC100m dataset, we saw an uneven distribution, with few images coming from countries in Africa and Asia. However, the locale of an image can be misleading, since if all the images taken in a particular country are only by tourists, this would not necessarily encompass the geo-representation one would hope for. Thus, here we combine our geography labels with language annotations. Fig.~\ref{fig:yf_lang} shows the percentage of images taken in a country and captioned in something other than the national language(s), as detected by the fastText library~\cite{joulin2016bag, joulin2016fasttext}. 
We use the lower bound of the binomial proportion confidence interval in the figure so that countries with only a few images total which happen to be mostly taken by tourists are not shown to be disproportionately imaged as so. Even with this lower bound, we see that many countries that are represented poorly in number are also under-represented by locals. To determine the implications in representation based on who is portraying a country, we categorize an image as taken by a local, tourist, or unknown, using a combination of language detected and tag content as an imperfect proxy. We then investigate if there are appearance differences in how locals and tourists portray a country by automatically running visual models. Although our tool does not find any such notable difference, this kind of analysis can be useful on other datasets where a local's perspective is dramatically different than that of a tourist's.

\subsubsection{Geography by Income}
\label{sec:geo_m_income}
Next, we consider how geography interacts with income. For this analysis, we focus on the portion of the BDD100K dataset in New York, and use income statistics by ZIP code~\cite{census_income, ny_income}. This dataset was collected by crowd-sourcing videos uploaded by drivers, a collection process that  has the potential to introduce geographic or socioeconomic biases due to the self-selection of drivers.

\begin{figure}
\centering
\includegraphics[width=0.49\textwidth]{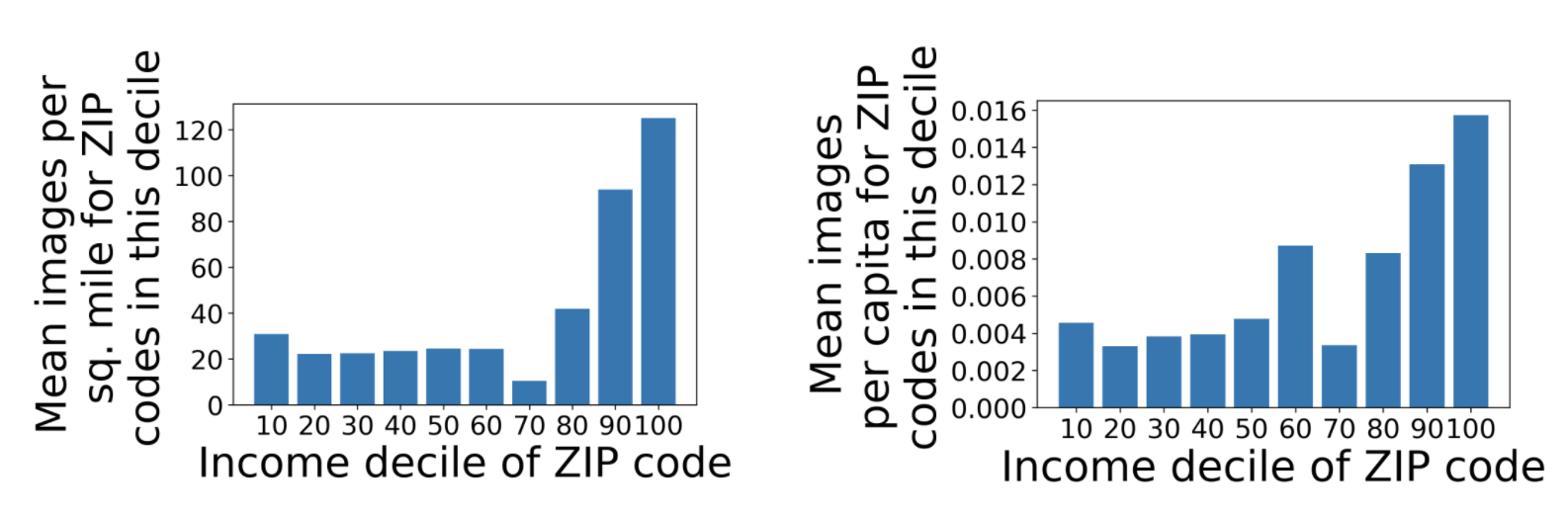}
\caption{ZIP codes with higher income are more represented in the BDD100K New York data.}
\label{fig:bdd_geo}
\end{figure}

To test whether this is the case, we divide the ZIP codes into deciles based on average income, and visualize how representation varies by income decline (Figure \ref{fig:bdd_geo}). We see that there is a large difference in the number of images per square mile between the two wealthiest deciles and the rest. It is possible that some of this may be explained by the wealthier ZIP codes being in boroughs with a greater density of roads. Accordingly, we also visualize the mean images per capita rather than per square mile, and find that a large difference persists. 

\begin{figure}
\centering
\includegraphics[width=0.49\textwidth]{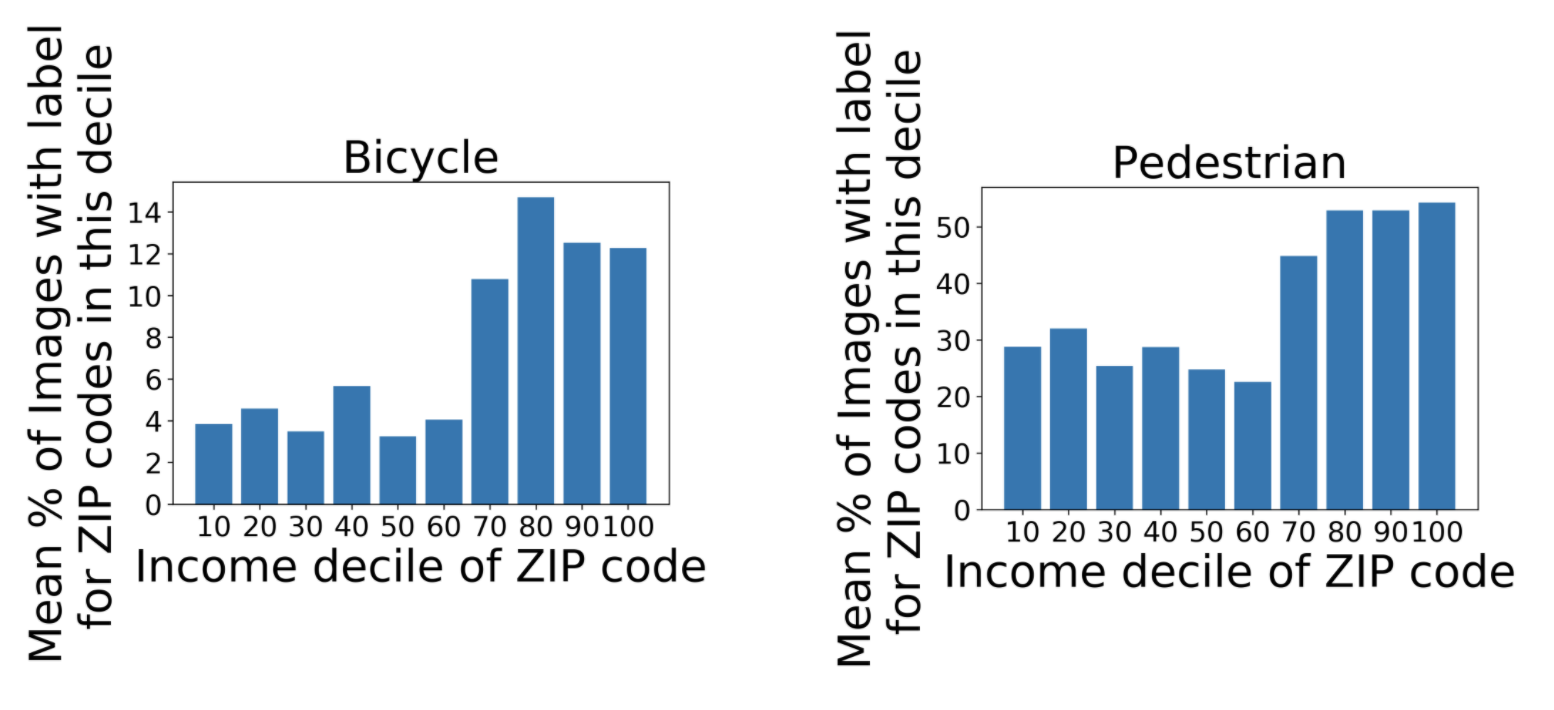}
\caption{ZIP codes with higher income are more likely to contain bicycles and pedestrians in the BDD100K New York data.}
\label{fig:bdd_tag}
\end{figure}

Such differences in representation can introduce biases or performance disparities in models trained on the data, because areas with different socioeconomic attributes are known to have systematic appearance differences~\cite{gebru2017street}. As evidence of such appearance differences in the BDD100K data, we highlight in Fig.~\ref{fig:bdd_tag} that income correlates with the presence of both the \obj{bicycle} and \obj{pedestrian} label.


\subsubsection{Geography by Weather}
\label{sec:geo_m_weather}
In the BDD100K self-driving car dataset, we have access to weather tags on each image. Weather is a very relevant factor for this context of automated driving, as oftentimes datasets only contain weather in clear conditions~\cite{sheeny2021radiate}, and thus have trouble generalizing to other weather conditions. Unsurprisingly, there are discrepancies between the weather distributions of images in the Northern California and New York City portions of this dataset, especially when looking at the \obj{snowy} label, which is present at 0.3\% for the former and 10\% for the latter. It is important to be aware of these differences when deploying models in a setting different from the one they were trained in. We note that while we have distinguished between geography analyses by object and by weather, both are automatically run through the same technical functionality of the tool, as they are considering how the variation of per-image tags, i.e., object and weather labels, vary by region.

\subsection{Geography-based Actionable Insights}
\label{sec:geo_actionable}
Much like the demographic-based actionable insights, those for geography-based are also more general and dependent on what the model trained on the data will be used for. Under- and over- representations can be approached in ways similar to before by augmenting the dataset, an important step in making sure we do not have a one-sided perspective of a region. Dataset users should validate that their models are not overfitting to a particular region's representation and image distribution by testing on more geographically diverse data, especially on that which is representative of where a model will be deployed. The geographic distribution of a dataset is intricately linked to the representations of objects and the people in them. Because of this, we note that not all instances of distribution differences are problematic and certain findings of the tool, such as an underrepresentation of \obj{safari} in the United States, may be entirely expected and not warrant any action to be taken. This will all depend on the use-case of the tested dataset.

It is clear that as we deploy more and more models into the world, there should be some form of either equal or equitable geo-representation. This emphasizes the need for data collection to explicitly seek out more diversity in locale, and specifically from the people that live there. Technology has been known to leave groups behind as it makes rapid advancements, and it is crucial that dataset representation does not follow this trend and base representation on digital availability. It requires more effort to seek out images from underrepresented areas, but as \textcite{Jo20} discuss, there are actions that can and should be taken, such as explicitly collecting data from underrepresented geographic regions, to ensure a more diverse representation.

\section{Discussion}
REVISE is effective at surfacing and helping mitigate many kinds of biases in visual datasets. But we make no claim that REVISE will identify \emph{all} visual biases. Creating an ``unbiased'' dataset may not be a realistic goal. The challenges are both practical (the sheer number of categories in modern datasets; the difficulty of gathering images from parts of the world where few people are online) and conceptual (how should we balance the goals of representing the world as it is and the world as we want it to be)?

The kind of interventions that can and should be performed in response to discovered biases will vary greatly depending on the dataset and applications. For example, for an object recognition benchmark, one may lean toward removing or obfuscating people that occur in images since the occurrence of people is largely incidental to the scientific goals of the dataset~\cite{Prabhu20, yang2021obfuscation}. But such an intervention wouldn't make sense for a dataset used as part of a self-driving vehicle application. Rather, when a dataset is used in a production setting, interventions should be guided by an understanding of the downstream harms that may occur in that specific application~\cite{barocas-hardt-narayanan}, such as poor performance in some neighborhoods. Making sense of which representations are more harmful for downstream applications may require additional data sources to help understand whether an underrepresentation is, for example, a result of a problem in the data collection effort, or simply representative of the world being imaged. Further, dataset bias mitigation is only one step, albeit an important one, in the much broader process of addressing fairness in the deployment of a machine learning system~\cite{green2018myth, birhane2021injustice}.

We also note that much of our analyses necessarily involves subdividing people along various socially-constructed dimensions. By operationalizing dynamic and non-discrete concepts such as gender and using skin tone as a proxy for race, we reify certain conceptions of these concepts~\cite{hanna20race, jacobs2021measurement} that harm certain groups, e.g., non-binary individuals~\cite{scheuerman2020gender, hamidi2018gender}.

\section{Conclusion}
In conclusion, we present the REVISE tool, which automates the discovery of potential biases in visual datasets and their annotations. We perform this investigation along three axes: object-based, person-based, and 

\vspace{-.3em}\noindent geography-based, and note that there are many more axes along which biases live. What cannot be automated is determining which of these biases are problematic and which are not, so we hope that by surfacing anomalous patterns as well as actionable next steps to the user, we can at least bring these biases to light.

\section{Acknowledgments}
This work is partially supported by the National Science Foundation under Grant No. 1763642 and No. 1704444. We would also like to thank Felix Yu, Vikram Ramaswamy, and Zhiwei Deng for their helpful comments, and Zeyu Wang, Deniz Oktay, and Nobline Yoo for testing out the tool and providing feedback.

\clearpage

\appendix
\section{Appendices}

\subsection{Gender label inference}
\label{app:per_m_infer}

\begin{figure}[t]
\centering
\includegraphics[height=2.67cm]{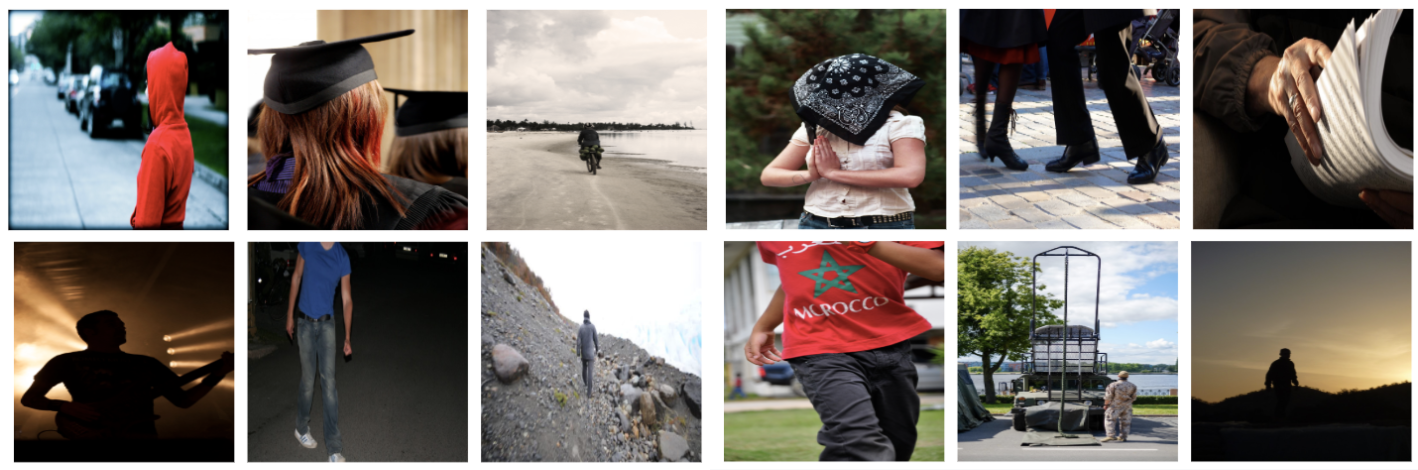}
\caption{Examples from OpenImages where annotators assigned gender to the person, but they should not have. The criteria used are that the person is either too small or has no face detected.}
\label{fig:1_oi_qual}
\end{figure}

An additional person-based metric we consider is gender label inference. Specifically, we note two especially concerning practices of assigning gender to a person who is too small to be identifiable, or no face is detected in the image.
This is not to say that if these cases are not met it is acceptable to assign gender, as gender cannot be visually perceived by an annotator, but merely that assigning gender when one of these two cases is applicable is a particularly egregious practice.
For example, it's been shown that in images where a person is fully clad with snowboarding equipment and a helmet, they are still labeled as male~\cite{Burns18} due to preconceived stereotypes. We investigate the contextual cues annotators rely on to assign gender, and consider the gender of a person unlikely to be identifiable if the person is too small (below 1000 pixels, which is the number of dimensions that humans require to perform certain recognition tasks in color images~\cite{Torralba08}) or if automated face detection (we used Amazon Rekognition~\cite{rekognition}, but note that any other face detection tool can be used) fails.
For COCO, we find that among images with a  human whose gender is unlikely to be identifiable, 77\% are labeled male. In OpenImages,\footnote{Random subset of size 100,000} this fraction is 69\%. Thus, annotators seem to default to labeling a person as male when they cannot identify the gender; the use of male-as-norm is a problematic practice~\cite{Moulton81}. Further, we find that annotators are most likely to default to male as a gender label in \obj{outdoor sports fields, parks} scenes, which is 2.9x the rate of female. Similarly, the rate for \obj{indoor transportation} scenes is 4.2x and \obj{outdoor transportation} is 4.5x, with the closest ratio being in \obj{shopping and dining}, where male is 1.2x as likely as female. This suggests that in the absence of gender cues from the person themselves, annotators make inferences based on image context. In Fig.~\ref{fig:1_oi_qual} we show examples from OpenImages where our tool determined that gender definitely should not be inferred, but was. Because attributes like skin tone can be inferred from parts of the image, such as a person's arm, we do not consider that attribute in this analysis.

This metric of gender label inference also brings up a larger question of which situations, if any, gender labels should ever be assigned~\cite{scheuerman2020gender, hamidi2018gender}. However, that is outside the scope of this work, where we simply recommend that dataset creators should give clearer guidance to annotators, and remove the gender labels on images where gender can definitely not be determined. We note that while we picked out two criteria of when a person is too small and when there is no face detected to be instances in which gender inference is particularly egregious, there are many other situations that users may wish to delineate for their own purposes.

\subsection{Validating Distance as a Proxy for Interaction}
\label{app:distance}
In Section~\ref{sec:gen_metrics}, Instance Counts and Distances, we make the claim that we can use distance between a person and an object as a proxy for if the person, $p$, is actually interacting with the object, $o$, as opposed to just appearing in the same image with it. This allows us to get more meaningful insight as to how genders may be interacting with objects differently. The distance measure we define is $dist = \frac{\textrm{distance between p and o centers}}{\sqrt{\textrm{area}_{\mathrm{p}}*\textrm{area}_{\mathrm{o}}}}$, which is a relative measure within each object class because it makes the assumption that all people are the same size, and all instances of an object are the same size. To validate the claim we are making, we look at the SpatialSense dataset~\cite{Yang19}; specifically, at 6 objects that we hope to be somewhat representative of the different ways people interact with objects: \obj{ball}, \obj{book}, \obj{car}, \obj{dog}, \obj{guitar}, and \obj{table}. These objects were picked over ones such as \obj{wall} or \obj{floor}, in which it is more ambiguous what counts as an interaction. We then hand-labeled the images where this object cooccurs with a human as ``yes" or ``no" based on whether the person of interest is interacting with the object or not. We pick the threshold by optimizing for mean per class accuracy, where every distance below it as classified as a ``yes" interaction and every distance above it as a ``no" interaction. The threshold is picked based on the same data that the accuracy is reported for. 

\begin{table}[]
\begin{center}
\begin{tabular}{ |>{\centering\arraybackslash}p{.75cm}|>{\centering\arraybackslash}p{.7cm}|>{\centering\arraybackslash}p{.7cm}|>{\centering\arraybackslash}p{1.5cm}|>{\centering\arraybackslash}p{1.5cm}|>{\centering\arraybackslash}p{.7cm}| } 
 \hline
 Object & \# Labeled Ex.'s & Mean Per Class Acc (\%) & ``Yes" Distance mean$\pm$std & ``No" Distance mean$\pm$std & Thre-shold\\ 
 \hline
 ball & 107 & 67 & $6.16\pm2.64$ & $8.54\pm4.15$ & 7.63\\ 
 \hline
 book & 27 & 78 & $2.45\pm1.99$ & $4.84\pm2.24$ & 3.88\\
 \hline
 car & 135 & 71 & $2.94\pm3.20$ & $4.59\pm2.97$ & 2.74\\
 \hline
 dog & 58 & 71 & $1.08\pm1.12$ & $2.07\pm1.79$ & 0.60\\
 \hline
 guitar & 40 & 88 & $0.90\pm1.77$ & $2.13\pm1.21$ & 1.61\\
 \hline
 table & 76 & 67 & $1.88\pm1.19$ & $3.28\pm2.45$ & 2.47\\
 \hline
\end{tabular}
\caption{\label{tbl:a_dist}Distances are classified as ``yes" or ``no" interaction based on a threshold optimized for mean per class accuracy. Visualization of the classification in Fig.~\ref{fig:a_dist}. Distances for ``yes" interactions are lower than ``no" interactions in all cases, in line with our claim that smaller distances are more likely to signify an interaction.}
\end{center}
\end{table}

\begin{figure}[t]
\begin{subfigure}{0.245\textwidth}
\includegraphics[width=\linewidth]{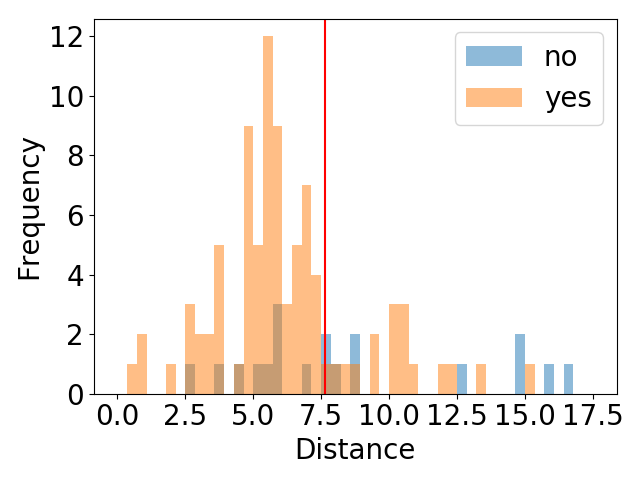}
\caption{Ball Distances} \label{fig:a}
\end{subfigure}\hspace*{\fill}
\begin{subfigure}{0.245\textwidth}
\includegraphics[width=\linewidth]{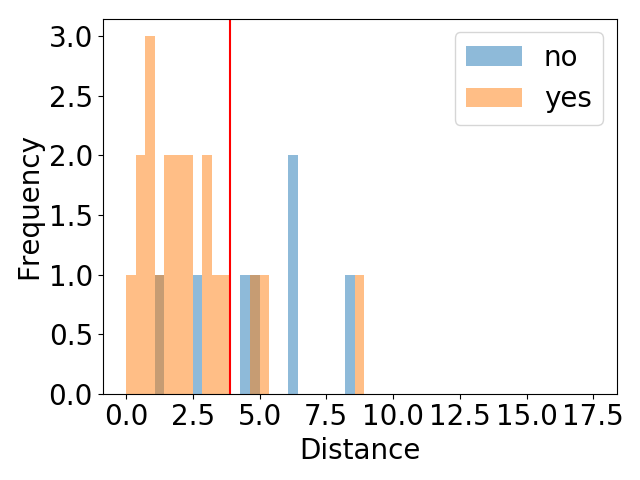}
\caption{Book Distances} \label{fig:b}
\end{subfigure}
\medskip
\begin{subfigure}{0.245\textwidth}
\includegraphics[width=\linewidth]{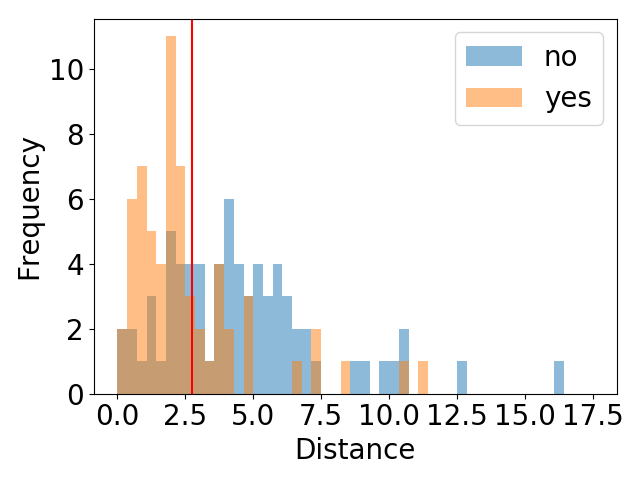}
\caption{Car Distances} \label{fig:c}
\end{subfigure}\hspace*{\fill}
\begin{subfigure}{0.245\textwidth}
\includegraphics[width=\linewidth]{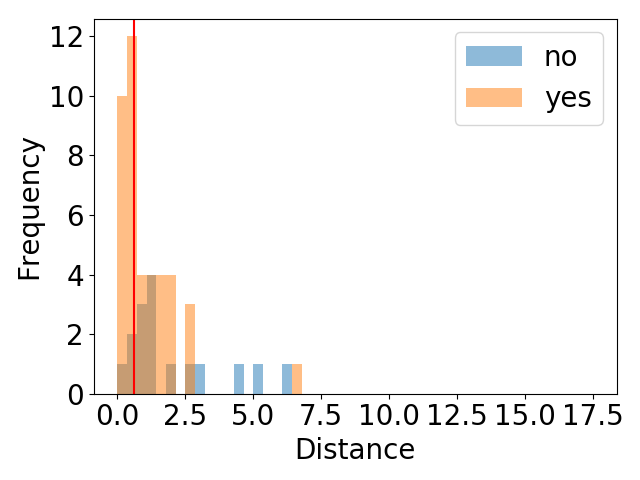}
\caption{Dog Distances} \label{fig:d}
\end{subfigure}
\medskip
\begin{subfigure}{0.245\textwidth}
\includegraphics[width=\linewidth]{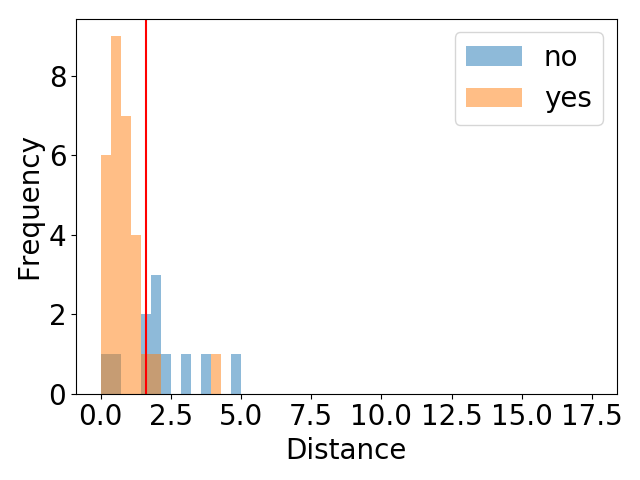}
\caption{Guitar Distances} \label{fig:e}
\end{subfigure}\hspace*{\fill}
\begin{subfigure}{0.245\textwidth}
\includegraphics[width=\linewidth]{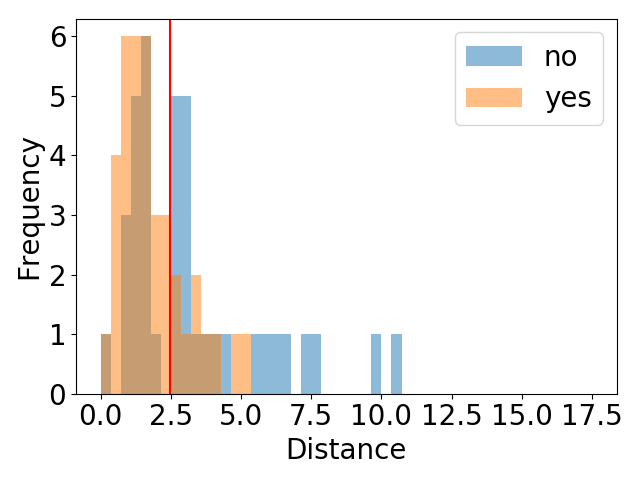}
\caption{Table Distances} \label{fig:f}
\end{subfigure}

\caption{Distances for the objects that were hand-labeled, orange if there is an interaction, and blue if there is not. The red vertical line is the threshold along which everything below is classified as ``yes", and everything above is classified as ``no".} \label{fig:a_dist}
\end{figure}

As can be seen in Table~\ref{tbl:a_dist}, for all 6 categories the mean of the distances when someone is interacting with an object is lower than that of when someone is not. This matches our claim that distance, although imperfect, can serve as a proxy for interaction. From looking at the visualization of the distribution of the distances in Fig.~\ref{fig:a_dist}, we can see that for certain objects like \obj{ball} and \obj{table}, which also have the lowest mean per class accuracy, there is more overlap between the distances for ``yes" interactions and ``no" interactions. Intuitively, this makes some sense, because a \obj{ball} is an object that can be interacted with both from a distance and from direct contact, and for \obj{table} in the labeled examples, people were often seated at a table but not directly interacting with it.

\subsection{Pairwise Queries}
\label{app:pairwise}
In Section~\ref{sec:obj_actionable}, another claim we make is that pairwise queries of the form ``[\obj{Desired Object}] and [\obj{Suggested Query Term}]" could allow dataset collectors to augment their dataset with the types of images they want. One of the examples we gave is that if one notices the images of \obj{airplane} in their dataset are overrepresented in the larger sizes, our tool would recommend they make the query ``\obj{airplane} and \obj{surfboard}" to augment their dataset, because based on the distribution of training samples, this combination is more likely than other kinds of queries to lead to images of smaller airplanes.

However, there are a few concerns with this approach. For one, certain queries might not return any search results. This is especially the case when the suggested query term is a scene category, such as \obj{indoor cultural}, in which the query ``\obj{pizza} and \obj{indoor cultural}" might not be very fruitful. To deal with this, we can substitute the scene category, \obj{indoor cultural}, for more specific scenes in that category, like \obj{classroom} and \obj{conference}, so that the query becomes something like ``\obj{pizza} and \obj{classroom}". When the suggested query term involves an object, there is another approach we can take. In datasets like PASCAL VOC~\cite{Everingham10}, the set of queries used to collect the dataset is given. For example, to get pictures of \obj{boat}, they also queried for \obj{barge}, \obj{ferry}, and \obj{canoe}. Thus, in addition to querying, for example, ``\obj{airplane} and \obj{boat}", one could also query for ``\obj{airplane} and \obj{ferry}", ``\obj{airplane} and \obj{barge}", etc.

Another concern is there might be a distribution difference between the correlation observed in the data and the correlation in images returned for queries. For example, just because \obj{cat} and \obj{dog} cooccur at a certain rate in the dataset, does not necessarily mean they cooccur at this same rate in search engine images. However, our query recommendation rests on the assumptions that datasets are constructed by querying a search engine, and that objects cooccur at roughly the same relative rates in the dataset as they do in query returns; for example, that because \obj{train} cooccurring with \obj{boat} in our dataset tends to be more likely to be small, in images returned from queries, \obj{train} is also likely to be smaller if \obj{boat} is in the image. We make an assumption that for an image that contains a \obj{train} and \obj{boat}, the query ``\obj{train} and \obj{boat}" would recover these kinds of images back, but it could be the case that the actual query used to find this image was ``coastal transit." If we had access to the actual query used to find each image, the conditional probability could then be calculated over the queries themselves rather than the object or scene cooccurrences. It is because we don't have these original queries that we use cooccurrences to serve as a proxy for recovering them.

\begin{figure}[t]
\centering
\begin{subfigure}{\textwidth}
    \includegraphics[height=6.9cm]{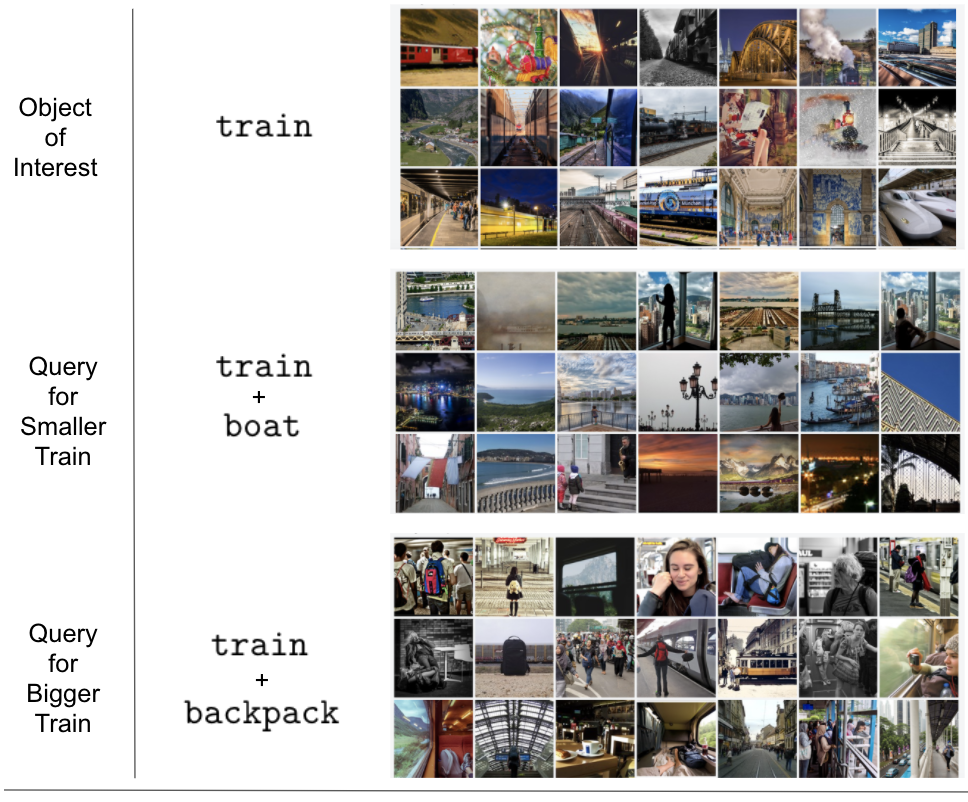}
\end{subfigure}
\begin{subfigure}{\textwidth}
    \includegraphics[height=6.9cm]{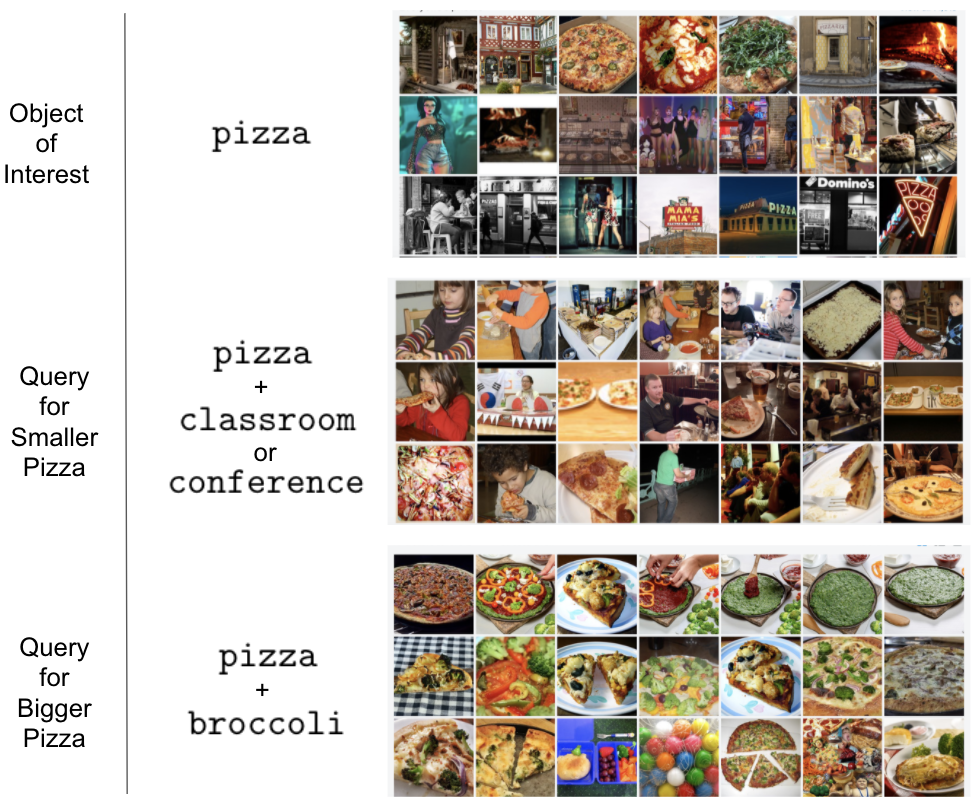}
\end{subfigure}
\caption{Screenshots of top results from performing queries on Flickr that satisfy the tags mentioned. For \obj{train}, when it is queried with \obj{boat}, the \obj{train} itself is more likely to be farther away, and thus smaller. When queried with \obj{backpack}, the image is more likely to show travelers right next to, or even inside of, a \obj{train}, and thus show it more in the foreground. The same idea applies for \obj{pizza} where it's imaged from further in the background when paired with an \obj{indoor cultural} scene, and up close with \obj{broccoli}.}
\label{fig:a_pairwise1}
\end{figure}

\begin{figure}[t]
\begin{subfigure}{\textwidth}
    \includegraphics[height=6.9cm]{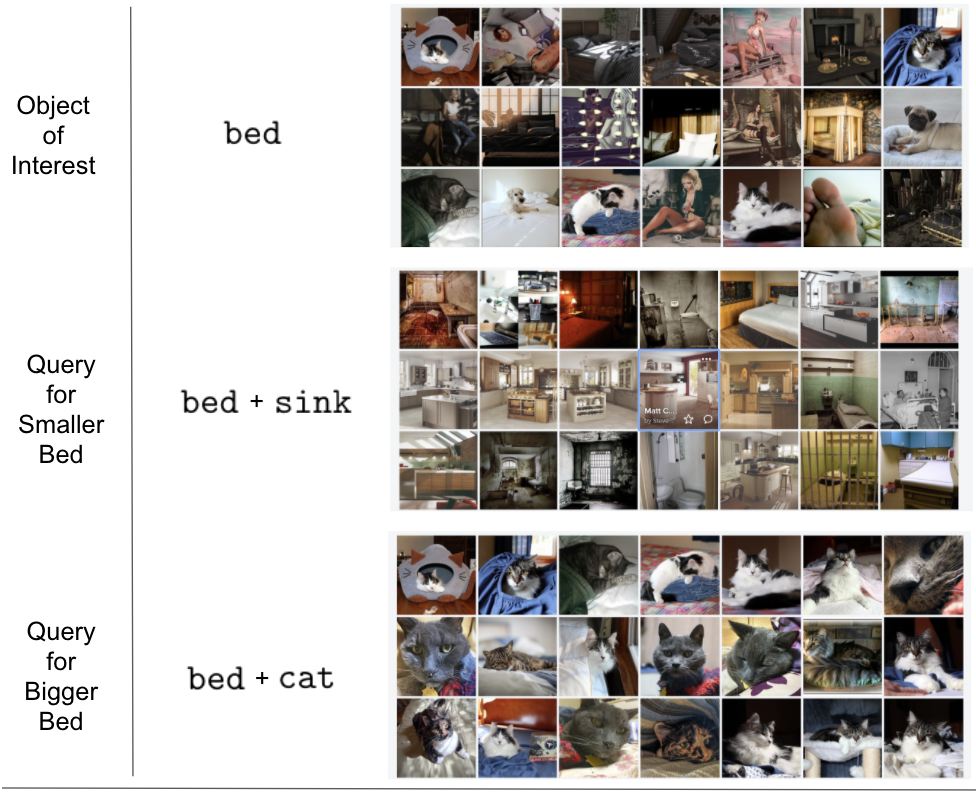}
\end{subfigure}
\begin{subfigure}{\textwidth}
    \includegraphics[height=6.9cm]{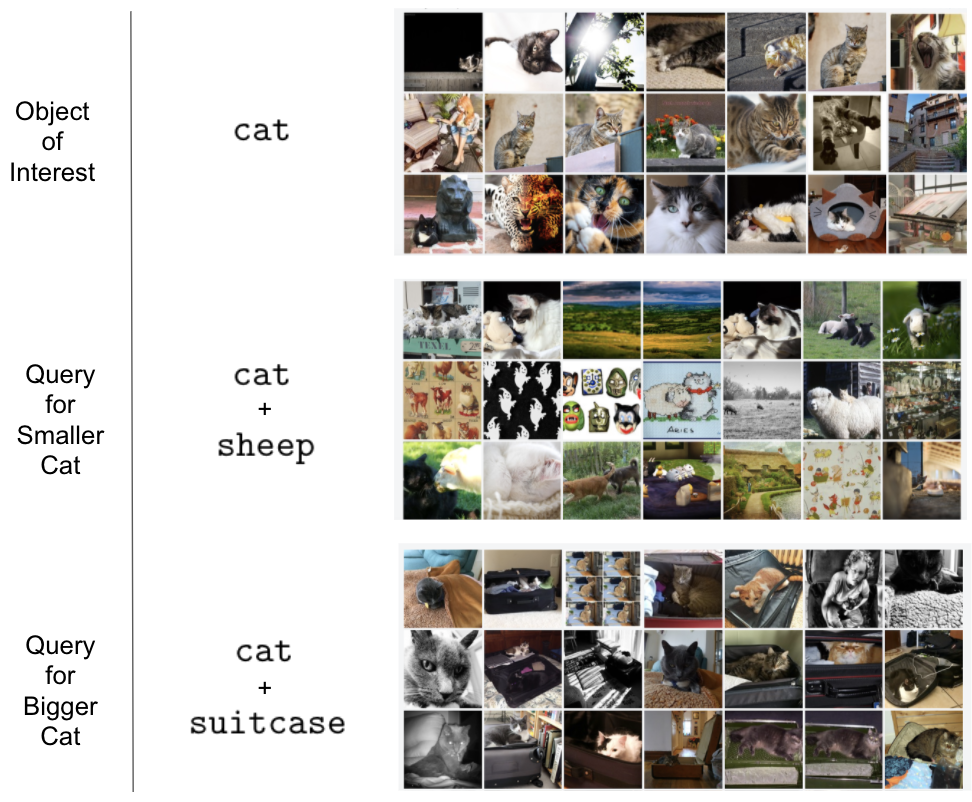}
\end{subfigure}
\caption{Screenshots of top results from performing queries on Flickr that satisfy the tags mentioned. For \obj{bed}, \obj{sink} provides a context that makes it more likely to be imaged further away, whereas \obj{cat} brings \obj{bed} to the forefront. The same is the case when the object of interest is now \obj{cat}, where a pairwise query with \obj{sheep} makes it more likely to be further, and \obj{suitcase} to be closer.}
\label{fig:a_pairwise2}
\end{figure}

To gain some confidence in our use of these pairwise queries in place of the original queries, we show qualitative examples of the results when searching on Flickr for images that contain the tags of the object(s) searched. We show the results of querying for (1) just the object (2) the object and query term that we would hope leads to more of the object in a smaller size, and (3) the object and query term that we would hope leads to more of the object in a bigger size. In Figs.~\ref{fig:a_pairwise1} and \ref{fig:a_pairwise2} we show the results of images sorted by relevance under the Creative Commons license. We can see that when we perform these pairwise queries, we do indeed have some level of control over the size of the object in the resulting images. For example, ``\obj{pizza} and \obj{classroom}" and ``\obj{pizza} and \obj{conference}" queries (scenes swapped in for \obj{indoor cultural}) return smaller pizzas than the ``\obj{pizza} and \obj{broccoli}" query, which tends to feature bigger pizzas that take up the whole image. This could of course create other representation issues such as a surplus of \obj{pizza} and \obj{broccoli} images, so it could be important to use more than one of the recommended queries our tool surfaces. Although this is an imperfect method, it is still a useful tactic we can use without having access to the actual queries used to create the dataset.\footnote{We also looked into using reverse image searches to recover the query, but the ``best guess labels" returned from these searches were not particularly useful, erring on both the side of being much too vague, such as returning ``sea" for any scene with water, or too specific, with the exact name and brand of one of the objects.}


\clearpage

\printbibliography

\end{document}